\documentclass[10pt,twocolumn,letterpaper]{article}

\usepackage{cvpr}              % To produce the CAMERA-READY version
\usepackage{comment}

\usepackage{adjustbox}
\usepackage{array}
\usepackage{multirow}
\usepackage{pifont}
\usepackage{multirow}
\usepackage[table,xcdraw]{xcolor} 
\usepackage[dvipsnames]{xcolor}
\usepackage{colortbl} 
\usepackage{algorithm}
\usepackage{algpseudocode}
\usepackage{amsmath} % For \text command
\usepackage{amssymb}
\usepackage{algorithm}       % For the floating 'algorithm' environment
\usepackage{physics}         % For \phase (defines e^{i...})
\definecolor{darkgreen}{RGB}{0,100,0}
\usepackage{enumitem}
 \usepackage{bm} 

\usepackage{pifont}% http://ctan.org/pkg/pifont

\definecolor{cvprblue}{rgb}{0.21,0.49,0.74}

\newcommand{\mytilde}{\raise.17ex\hbox{$\scriptstyle\mathtt{\sim}$}}

\usepackage{xspace}

\newcommand{\methodname}{PackUV\xspace}
\newcommand{\fittername}{PackUV-GS\xspace}
\newcommand{\numimgs}{2B}
\newcommand{\numsequences}{100\xspace}
\newcommand{\datasetname}{\methodname-\numimgs\xspace}

\newcommand{\phase}[1]{\textsc{#1}}
\newcommand{\varr}[1]{\texttt{#1}}    % Variable names in monospace

\definecolor{cvprblue}{rgb}{0.21,0.49,0.74}
\usepackage[pagebackref,breaklinks,colorlinks,allcolors=cvprblue]{hyperref}

\def\paperID{21244} % *** Enter the Paper ID here
\def\confName{CVPR}
\def\confYear{2026}

\title{\methodname: Packed Gaussian UV Maps for 4D Volumetric Video}

\author{
Aashish Rai$^{1}$
\and
Angela Xing$^{1}$
\and
Anushka Agarwal$^{2}$
\and
Xiaoyan Cong$^{1}$
\and
Zekun Li$^{1}$
\and
Tao Lu$^1$
\and
Aayush Prakash$^3$
\and
Srinath Sridhar$^2$
\vspace{0.05in}
\and
\centerline{$^1$Brown University\hspace{0.2in} $^2$UMass Amherst \hspace{0.2in} $^3$Meta}
\vspace{0.2cm}
\and
{\tt\small \url{https://ivl.cs.brown.edu/packuv}}
}

\begin{document}
% \vspace{-0.9cm}
% \vspace{-0.8cm}
\twocolumn[{%
\renewcommand\twocolumn[1][]{#1}%
\maketitle
\vspace{-0.3cm}
\centering
    %\hspace*{0.5cm}
    \includegraphics[width=\textwidth]{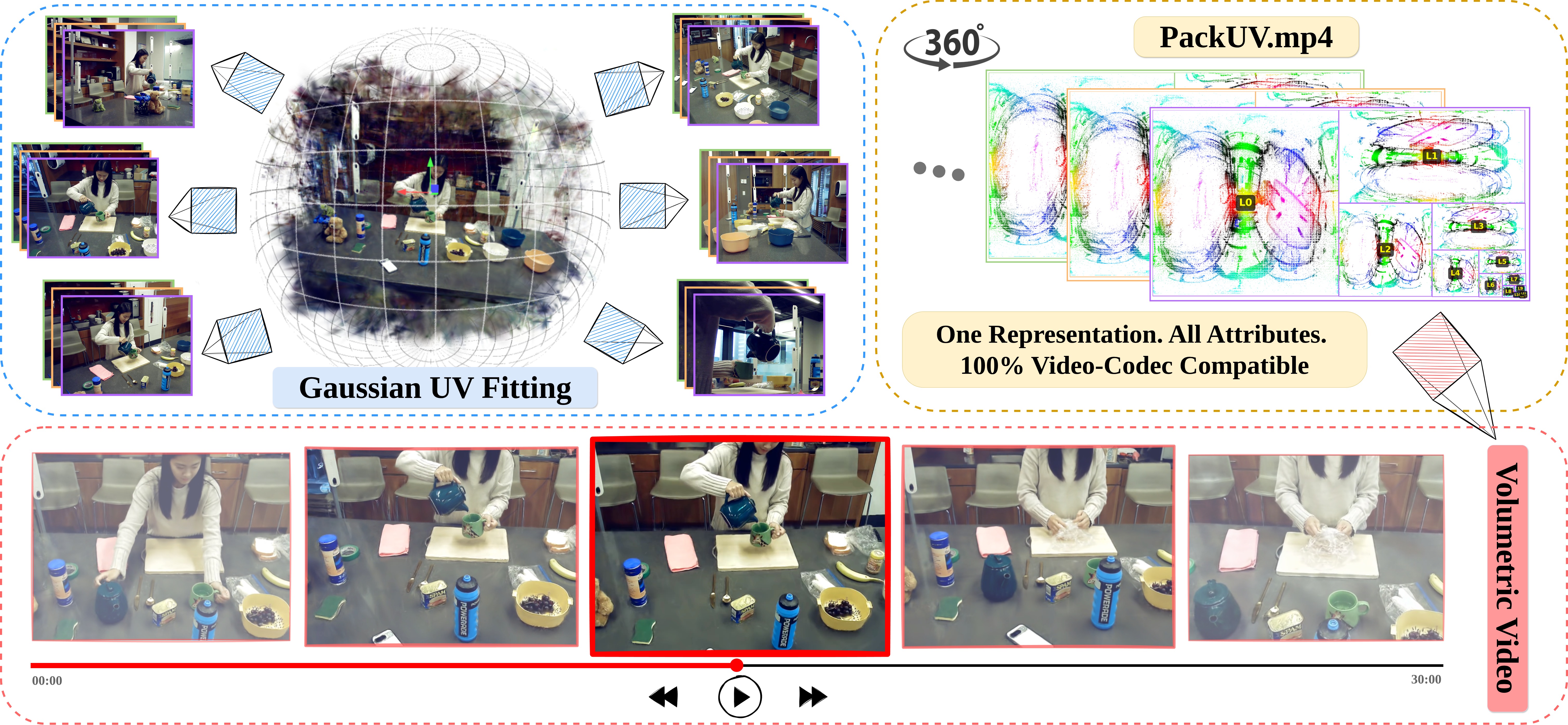}
    % \vspace*{-//0.2cm}
    \captionof{figure}{\textit{
    We propose a novel and compact 4D representation, \textbf{\methodname}, for volumetric videos that packs 3D Gaussian attributes into a sequence of 2D UV atlases (\textcolor{Dandelion}{yellow}, top right).
    \methodname is readily compatible with existing video coding infrastructure (\eg,~can be coded with HEVC, FFV1).
    We also propose \textbf{\fittername}, a method to directly fit Gaussian attributes from multi-view RGB videos into structured \methodname (\textcolor{cyan}{blue}, top left) via optical flow-guided keyframing and Gaussian labeling to fit arbitrary length sequences with temporal consistency even in the presence of large motions and disocclusions. 
    The fitted scene can be rendered back to streamable volumetric video from any viewpoint (\textcolor{Salmon}{red}, bottom).
    We also propose \datasetname, the largest 4D multi-view dataset containing \textbf{\numimgs} frames
    captured with over 50 synchronized cameras to provide 360$^\circ$ coverage.
    }}
    \vspace{4mm}
    \label{fig:teaser}
}]

\maketitle

\begin{abstract}
% \vspace{-6mm}

Volumetric videos offer immersive 4D experiences, but remain difficult to reconstruct, store, and stream at scale. 
Existing Gaussian Splatting based methods achieve high-quality reconstruction but break down on long sequences, temporal inconsistency, and fail under large motions and disocclusions. 
Moreover, their outputs are typically incompatible with conventional video coding pipelines, preventing practical applications. 
We introduce PackUV, a novel 4D Gaussian representation that maps all Gaussian attributes into a sequence of structured, multi-scale UV atlases, enabling compact, image-native storage. 
To fit this representation from multi-view videos, we propose \fittername, a temporally consistent fitting method that directly optimizes Gaussian parameters in the UV domain. 
A flow-guided Gaussian labeling and video keyframing module identifies dynamic Gaussians, stabilizes static regions, and preserves temporal coherence even under large motions and disocclusions. 
The resulting UV atlas format is the first unified volumetric video representation compatible with standard video codecs (e.g.,~FFV1) without losing quality, enabling efficient streaming within existing multimedia infrastructure. To evaluate long-duration volumetric capture, we present PackUV-2B, the largest multi-view 4D dataset to date, featuring more than 50 synchronized cameras, substantial motion, and frequent disocclusions across \numsequences sequences and \numimgs~(billion) frames. Extensive experiments demonstrate that our method surpasses existing baselines in rendering fidelity while scaling to sequences up to 30 minutes with consistent quality.

\vspace{-2mm}

\end{abstract}
    
% \vspace{-1mm}
\section{Introduction}
\label{sec:intro}

% \vspace{-1mm}
Volumetric videos are a form of immersive media that capture scenes in three dimensions and across time (4D), making them viewable from any perspective.
% They enable interactive 6DoF viewing and immersive experiences in 
They promise numerous applications in AR/VR, entertainment, sports, as well as in applications requiring 4D understanding, for instance, in robotics~\cite{YOUNG2023591,smolic2022volumetric,volumetricvideoXR,DVB_S101_2024}. 
Unsurprisingly, creating volumetric videos from multiple camera views has been a long-standing challenge in computer vision and graphics~\cite{meshry2019neural, thies2018ignor, bansal20204d, cohen1996the, Sohn2004}. 

Common approaches for volumetric video reconstruction rely on explicit representations, such as point clouds~\cite{reimat2021cwipc,sun2023dynamic,hu2023fsvvd} meshes~\cite{pages2021volograms}, multi-plane images~\cite{srinivasan2019pushing, tucker2020single,chen2022casual,WANG2020247}, or multi-sphere images~\cite{attal2020matryodshka, broxton2020immersive}. However, these methods are limited in their ability to render complex scenes and are highly memory intensive.
% \tao{move these old work to related work.}
Meanwhile, radiance fields~\cite{mildenhall2020nerf,xie2022neural} and 3D Gaussian Splatting ~\cite{kerbl3Dgaussians} 
% \aayush{why not mention 4DGS directly, is there a reason we are avoiding it?} 
% \rai{4DGS is one specific type of FVV method.}
have emerged as leading representations for 3D reconstruction and volumetric video~\cite{sun20243dgstream, wu20244d,yang2023gs4d,yang2024deformable,chen2025adaptive,lee2024compact}. 
% However, methods struggle to operate on videos longer than a few seconds~\cite{wu20244d,liang2024gaufregaussiandeformationfields,xu20244k4d,yang2024deformable,yang2023gs4d,chen2025adaptive}. 
Despite their success, these methods~\cite{wu20244d,liang2024gaufregaussiandeformationfields,xu20244k4d,yang2024deformable,yang2023gs4d,chen2025adaptive} struggle to operate on videos longer than a few seconds.
% \anush{"also, we have already mentioned long duration videos/ vidoes longer than view seconds above, do we need to remention it here?}
\emph{Streaming} approaches~\cite{sun20243dgstream,luiten2024dynamic,xu2024longvolcap,lee2024compact} address this via online fitting but struggle to retain \textbf{long-duration temporal consistency}, capture \textbf{large motions}, and handle \textbf{disocclusions} (\eg,~when a new object enters the scene).
% \tao{optical flow is a local computation across short timespan, no need to highlight this part in the beginning and with bold font.} \axing{I think this is ok because we are using temporal consistency to pick keyframes so it operates on the whole sequence}
Furthermore, volumetric videos produced by these methods cannot be seamlessly shared due to their large size and required bespoke compression methods~\cite{dai20254d_motion_layering} that are incompatible with existing multimedia infrastructure. 
% \tao{only this sentence is highly related to the title/our core contribution, move it forward.}\tao{analyze some limitations of UV-GS here.}

To address these challenges, we introduce \textbf{\methodname}, a novel 4D Gaussian representation that packs 3D Gaussian attributes into a sequence of structured, multi-layered 2D UV maps. 
These UV map layers are further compacted into a single progressive atlas (\Cref{fig:teaser}), enabling efficient storage.
We also present \textbf{\fittername}, a fitting method that generates temporally-consistent volumetric videos from multi-view input. 
Unlike previous 3DGS-to-2D approaches that rely on lossy post-hoc UV unwrapping~\cite{rai2025uvgs, xiang2025repurposing}, \fittername directly fits Gaussian parameters into the UV domain. 
To maintain temporal coherence under large motions and disocclusions, an optical-flow–guided module identifies dynamic Gaussians, enforces flow-based keyframing, and selectively freezes gradients in static regions to ensure stable optimization.
This design supports high-quality reconstruction for sequences of arbitrary duration while maintaining quality.
% To solve these challenges, we propose a novel 4D Gaussian representation, \textbf{\methodname}, which stores the Gaussian attributes into a structured, multi-layer 2D UV domain.
% The multi-layer 2D UV maps are then reorganized and packed as a single layer progressive atlas map(Fig.~\ref{fig:teaser}).
% We further introduce a method, \textbf{\fittername}, to generate temporally consistent volumetric videos from multi-view input.
% Unlike prior 3DGS to 2D mapping approaches that require lossy post-optimization UV mapping~\cite{rai2025uvgs, xiang2025repurposing}, 
% \fittername directly fits 3D Gaussian attributes into the UV domain.
% Our method supports high quality reconstruction for scenes of arbitrary duration.
% To retain temporal consistency while accommodating large motions and disocclusions, an optical-flow based module identifies dynamic Gaussians, enforces coherence via flow-guided keyframing, and selectively freezes gradients to stabilize static regions. 
The resulting image-native representation replaces point-centric storage with a sequence of ordered, multi-scale 2D UV atlases, enabling efficient streaming and storage via standard video coding methods (\eg,~HEVC, FFV1). 
To the best of our knowledge, this is the 
\emph{first unified representation that applies conventional video coding directly to all 3DGS attributes}, 
with no quality loss during the conversion, to bridge 4D Gaussian representations and the existing video infrastructure.

We evaluate the performance of our method and justify design choices on a variety of existing datasets.
% However, existing datasets~\cite{Joo_2015_ICCV,li2022neural,yan2023nerf,xu2024representing,pumarola2020d} focus primarily on scenes captured using only frontal cameras rather than 360$^\circ$, have small motions, and don't contain disocclusions.  
However, existing datasets~\cite{Joo_2015_ICCV,li2022neural,yan2023nerf,xu2024representing,pumarola2020d} are largely restricted to frontal cameras and exhibit limited motions and  disocclusions.
To better showcase the abilities of our representation and compare it with existing work, we captured the largest long-duration multi-view video dataset, \textbf{\datasetname}.
% Unlike most existing datasets with only front-facing cameras, 
\datasetname features real-world dynamic scenes with more than 50 synchronized cameras, providing 360$^\circ$ coverage in both controlled studio and uncontrolled in-the-wild settings. 
% \anush{We mentioned frontal facing above}
% In total, our dataset contains \textbf{\numsequences sequences} with more than \textbf{\numimgs (billion) frames}.
% Our dataset includes various scenarios, like human-human interaction, human-object interaction, human-robot interaction, and many more.   
In total, \datasetname contains \numsequences sequences with more than \textbf{\numimgs~(billion) frames} featuring a diverse range of scenarios including human-human, human-object, and human-robot interactions.
% , and many more.
% This presents further diversities and challenges to the 4D reconstruction task. 
% \anush{These variations pose significant challenges for 4D reconstruction, highlighting the need for robust and temporally consistent representations.}
%, hoping to inspire new perspectives and questions.
Extensive experiments show that our method outperforms all baselines across standard metrics and can model much longer sequences (up to 30 minutes) while maintaining consistent quality.
To summarize, our contributions include:
\begin{itemize}[noitemsep, topsep=0pt, parsep=0pt, partopsep=0pt]

    \item \textbf{\methodname}, a new volumetric video representation that packs 3D Gaussian attributes into a sequence of UV atlases for efficient streaming and storage, making it readily compatible with existing video coding infrastructure.
    \item \textbf{\fittername}, an efficient method to fit \methodname directly from multiview videos using optical-flow-based keyframing and Gaussian labeling to handle large motions, disocclusions, and temporal consistency. 
    \item \textbf{\datasetname}, the largest multi-view 4D dataset with \numimgs~frames, large motions, and disocclusions.
    It provides 360$^\circ$ coverage from 50+ synchronized cameras.
\end{itemize}

\section{Related Work}
\label{sec:related_works}
In this brief related work, we focus on methods and representations for reconstructing volumetric videos.

% \vspace{-4mm}

\paragraph{4D Volumetric Video.}
Volumetric video fitting is a long-standing problem, with works originally focusing on using 
% aim to offer users immersive interaction experiences with the 4D world, which is long-standing research problem in computer vision and graphics.
% Various methods have have been proposed, including 
multi-view images~\cite{10.1145/383259.383309, 10.1145/2487228.2487238, 10.1145/3130800.3130855, 10.1145/1015706.1015766}, multi-plane images~\cite{srinivasan2019pushing, tucker2020single,chen2022casual,WANG2020247}, light fields~\cite{10.1145/2980179.2980251, davis2012unstructured, 10.1145/3596711.3596760, 10.1145/3596711.3596759} as well as explicit representations like point clouds~\cite{reimat2021cwipc,sun2023dynamic,hu2023fsvvd} meshes~\cite{pages2021volograms}, voxels~\cite{10.1145/2897824.2925969, 10.1145/54852.378484, 7298631, 10.1145/2984511.2984517}, or multi-sphere images~\cite{attal2020matryodshka,broxton2020immersive}. 
More recently, radiance fields~\cite{mildenhall2020nerf, barron2021mip, chen2022tensorf}, in particular 3D Gaussian splatting~\cite{kerbl3Dgaussians}, have emerged as the de facto method for static novel view synthesis and reconstruction.
% NeRF~\cite{mildenhall2020nerf} represents the scene by mapping the 3D coordinate and view direction to the color and opacity, producing photorealistic renderings through differentiable volume rendering.
% In particular, 3D Gaussian Splatting (3DGS)~\cite{kerbl3Dgaussians} has become the  achieved photorealistic novel view synthesis with rapid training speeds and real-time rendering, emerging as a compelling alternative for static scene modeling. 
Building on 3DGS, a significant amount of work on dynamic scenes has subsequently emerged~\cite{huang2024sc, kratimenos2025dynmf, lin2024gaussian, luiten2023dynamic, yang2024deformable, wu20244d, xie2024physgaussianphysicsintegrated3dgaussians, yang2023real, duan20244d, yan20244d, chen2025adaptive,gan2023v4d, song2023nerfplayer, xu20244k4d, li2022streaming, pumarola2021d, 2025dytact, attal2023hyperreel, icsik2023humanrf, kim2024sync, lin2023high, peng2023representing, wang2022fourier, pokhariya2024manus, wang2023neural, li2022neural, shao2023tensor4d, fridovich2023k, cao2023hexplane, wang2023mixed}.
Deformable3DGS~\cite{yang2024deformable}, 4DGS~\cite{wu20244d}, and Grid4D~\cite{xu2024grid4d} define a deformation field mapping canonical Gaussian primitives to specific time steps.
RealTime4DGS~\cite{yang2023real} and 4D-Rotor-GS~\cite{duan20244d} introduce 4D Gaussian primitives, improving flexibility for a variety of dynamic scenes.
% Dynamic3DGS~\cite{luiten2023dynamic} learns transformations of Gaussian primitives over time, facilitating dynamic tracking~\cite{wang2023tracking}. 
% Deformable3DGS~\cite{yang2024deformable}, 4DGS~\cite{wu20244d}, and Grid4D~\cite{xu2024grid4d} define a deformation field mapping canonical Gaussian primitives to specific time steps.
% RealTime4DGS~\cite{yang2023real} and 4D-Rotor-GS~\cite{duan20244d} introduce 4D Gaussian primitives, improving flexibility for a variety of dynamic scenes.
% Numerous methods extend NeRF or 3DGS to dynamic settings~\cite{gan2023v4d, song2023nerfplayer, xu20244k4d, li2022streaming, pumarola2021d, attal2023hyperreel, icsik2023humanrf, kim2024sync, lin2023high, peng2023representing, wang2022fourier, wang2023neural, li2022neural, shao2023tensor4d, fridovich2023k, cao2023hexplane, wang2023mixed}.
% DyNeRF~\cite{li2022neural} leverages time-conditioned latent codes to model dynamics. 
% Tensor4D~\cite{shao2023tensor4d}, K-Planes~\cite{fridovich2023k}, and HexPlane~\cite{cao2023hexplane} exploit planar factorization in the 4D spacetime domain. 
% MixVoxels~\cite{wang2023mixed} adopts a mixture of static and dynamic voxel representations for faster rendering. 
Despite the progress, these methods are limited to short sequences (few seconds), memory intensive, temporally inconsistent, or cannot handle disocclusions (see \Cref{sec:experiments}).
% However, these methods are memory intensive and can only handle short dynamic scenes around $1\sim2$ seconds at $30$ FPS.
% most of these methods require extensive sampling, that can hinder real-time rendering~\cite{barron2021mip, chen2022tensorf, mildenhall2020nerf}.

% \aayush{we seriously need to cut the related work by half. You can do that by 1/ cutting down method details of 4DGS methods- focus on only details that help differentiate PackUV-GS from them and not describing these methods in any details 2/ Cut down details on 3DGS, or first paragraph}

% However, extending to longer ``time'' dimension is non-trivial due to memory inefficiency~\cite{yang2024deformable,chen2025adaptive,Wu_2024_CVPR,yang2023gs4d} and temporal inconsistency~\cite{sun20243dgstream,luiten2024dynamic}.
% However, these methods are memory intensive and can only handle short dynamic scenes around $1\sim2$ seconds at $30$ FPS.
% \axing{like abstract, needs verification} \rai{yes, for classical deformation based methods.}
Recent methods like LongVolCap~\cite{xu2024representing} make tremendous progress by leveraging a hierarchical temporal 4D Gaussian representation to compactly model long-horizon scenes, but still struggle to fit arbitrary durations due to the growth of Gaussians.
% \srinath{Where is the motion layering paper?}
% because the number of Gaussians during training is associated with the length of volumetric videos.
% To address the memory cost of offline 3DGS fitting, 3DGStream~\cite{sun20243dgstream} proposes an online training manner to learn per-frame 3D Gaussians with a neural transformation cache to model the dynamics, which effectively reduces the storage cost and improves the training efficiency.
To address the memory cost of offline 3DGS fitting, 3DGStream~\cite{sun20243dgstream} and ATGS~\cite{chen2025adaptive} propose online training and model dynamics of 3D Gaussians per-frame with a neural transformation cache. 
However, the per-frame storage cost of 3D Gaussians remains large, making it infeasible to transmit and play in a streaming manner like conventional videos. 
% Importantly, the per-frame storage cost of 3D Gaussians remains prohibitively large, making it infeasible to transmit and play in a streaming manner like conventional videos.
% However, it still suffers from flickering artifacts and struggles to model the emerging objects.
% \rai{discuss Ex4DGS, Grid4D, Motion layering, GIFStream}
Ex4DGS~\cite{lee2024fully} addresses memory overhead by separating Gaussians into linearly moving `static' and fully dynamic Gaussians.
In addition to static-dynamic Gaussian separation, GIFstream~\cite{Li_2025_CVPR} and Motion Layering~\cite{dai20254d_motion_layering} improve dynamic modeling through time-dependent feature streams.
However, these methods still struggle to model large motions and disocclusions.
% Other methods like Grid4D~\todo{cite grid4d} use a deformable multi-channel hash-grid representation that separately models spatial and temporal hash encodings through a tri-axial 4D decomposition, thereby eliminating reliance on low-rank plane-based assumptions. However, it remains limited in its ability to tackle complex and long range motion. Similarly, GIFstream \todo{cite gif stream} and Motion layering \todo{cite motion layering} attempt to improve dynamic modeling through time-dependent feature streams and static–dynamic separation, respectively, but are quite restricted in their ability to incorporate fast moving objects or large-magnitude motion. Ex4DGS~\todo{cite ex4dgs} on the other hand aims toward overcoming computational overload by avoiding heavy neural deformation networks and instead representing motion explicitly via sparse keyframes. However, the explicit formulation still struggles with newly appearing or disappearing objects in a complex motion frame. 

\vspace{-4mm}
\paragraph{Volumetric Video Representations.}
%
% Representing large-scale 3D Gaussian-based scene reconstruction often involves millions of Gaussians, demanding several gigabytes of storage for real-world applications. 
% To address this limitation, recent works have explored various compression strategies. 
% For instance, Compact3D~\cite{lee2024compact} and CompGS~\cite{navaneet2024compgs} adopt vector quantization to store Gaussians in compact codebooks.
% In parallel, since the 3DGS fitting results usually suffer from high redundancy, pruning-based approaches~\cite{ali2024trimming, fan2024lightgaussian, fang2024mini, liu2025efficientgs, niemeyer2024radsplat, papantonakis2024reducing} propose pruning schemes to remove a portion of Gaussians based on predefined importance criteria.
% \aayush{Cut down details in this section to describe each related work, focus on differentiators only}
Representing and storing volumetric videos is significantly more expensive than 2D images or videos due to the additional spatial and temporal dimensions.
Several approaches have been proposed, either by pruning unused Gaussians, cleverly compressing Gaussian attributes, or using learning~\cite{,lee2024compact, navaneet2024compgs, ali2024trimming, fan2024lightgaussian, fang2024mini, liu2025efficientgs, niemeyer2024radsplat, papantonakis2024reducing}.
Recently, representations for `flattening' 3D Gaussians into 2D form have been receiving attention, for instance, SOG~\cite{morgenstern2024sog} and UV projection~\cite{xiang2025repurposing,rai2025uvgs}.
However, all these methods are limited to static 3D scenes.
For 4D volumetric videos a naive sequence of static representations would still be too large.

% Additionally, only based on pruning the number of 3D Gaussians for each frame, it is still challenging to achieve the real-time 4D live stream ($\sim$ 30 FPS) with consumer grade infrastructure.
% Hence, most of the previous methods are incompatible with the widely available video coding infrastructure. 
Recognizing this, some recent works~\cite{dai20254d_motion_layering, vcubed} transfer ideas from 2D video coding to volumetric videos.
However, due to the unstructured nature of regular 3DGS attributes, these methods have relatively large compression losses, or heavy computational requirements.
% they can only apply the video coding to a few 3DGS attributes.
% The unstructured 3D Gaussian points-based representation makes it hard to leverage the temporal redundancy to compress the 3D Gaussian sequence further. 
Structured UV projection methods~\cite{rai2025uvgs,xiang2025repurposing} are promising since they can then be combined with existing image or video coding methods.
However, projecting an already-optimized 3DGS into a UV map is lossy and computationally redundant.
% , the post-optimization mapping strategy of UVGS is very lossy when applying to real-world scenes due to the significantly large number of Gaussian primitives and limited budget in the UV space.

Our \methodname representation, together with our native \fittername fitting method, overcomes these limitations by directly producing a sequence of UV atlases that are fully compatible with existing lossless (and lossy) video coding methods (\eg,~HEVC, AVC, FFV1).
In addition, our method can handle arbitrary length sequences while preserving temporal consistency under large motions and disocclusions.
% formats like h265, and the advances in 2D mapping of 3D attributes~\cite{rai2025uvgs, xiang2025repurposing}, we propose a novel lossless image-based representation 
% \methodname to organize the 3D Gaussian in structural multi-scale UV images, that can easily use the readily-available video coding infrastructure for 4D streaming. 
% We hope our method makes the streaming of volumetric videos possible in real-time for various downstream applications\cite{4dreport}.

\begin{figure*}[!ht]
\centering
% \vspace{-0.8cm}
\includegraphics[width=6.8in]{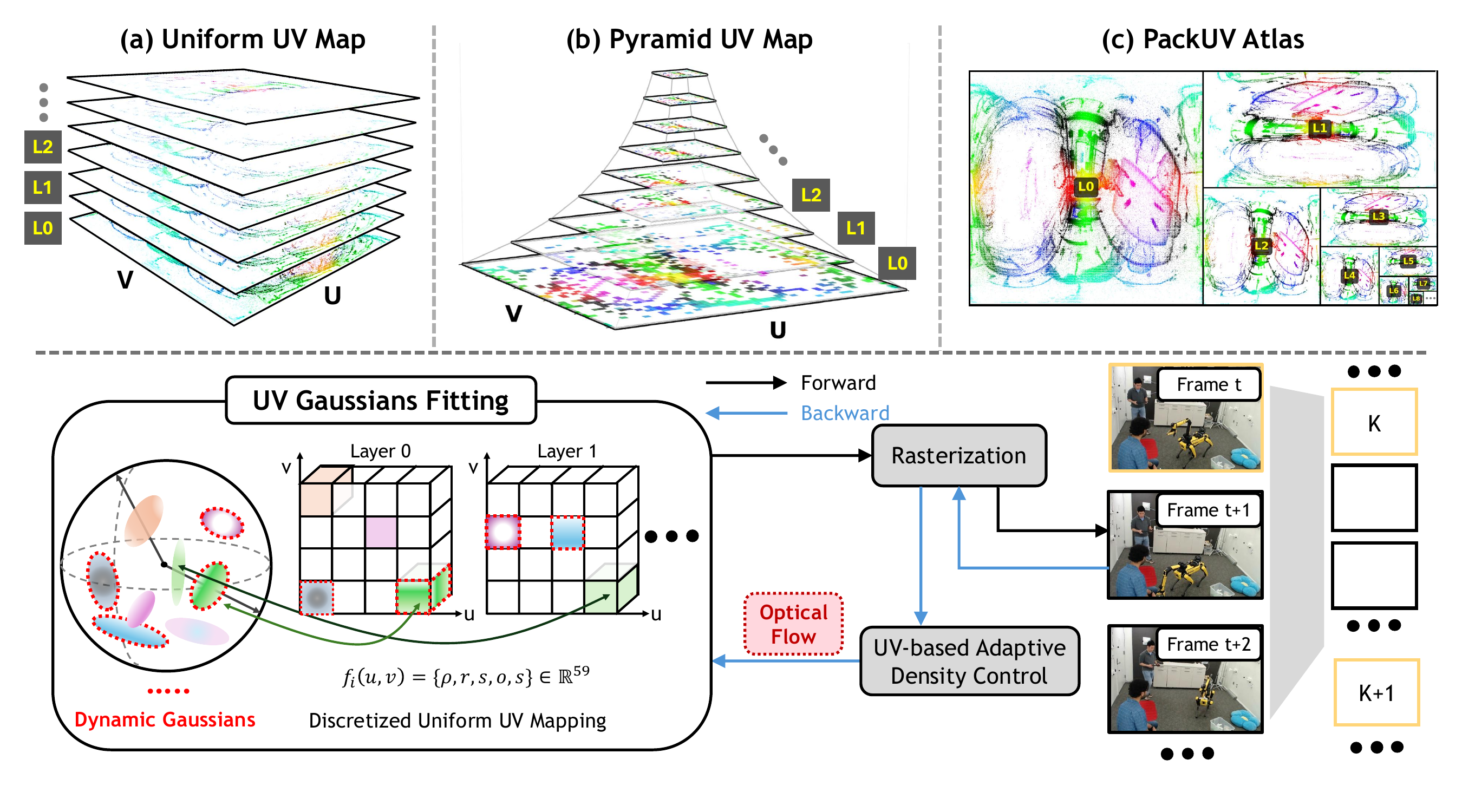} 
\vspace{-0.4cm}
\caption{
(Top) Three UV-map organization strategies:
(a) naïvely stacking UV layers (deep layers become more and more sparse);
(b) a geometric-progression UV pyramid (more uniform sparsity with less storage);
(c) \methodname, which packs all pyramid layers into a single UV atlas for efficient, codec-friendly processing.
% (Top) Shows three strategies for organizing UV maps.
% (a) A straightforward stack of $K$ UV layers storing Gaussian primitives; however, deeper layers become increasingly sparse.
% (b) A UV pyramid that allocates layer resolution using a geometric progression, maintaining more uniform sparsity while reducing storage.
% (c) \methodname - a reorganized version of the pyramid in which all layers are packed into a single UV atlas, enabling efficient processing and codec-friendly storage.
(Bottom) We propose \fittername, a new representation based on 3DGS with a discrete spatial distribution constraint via UV fitting. 
It uses multiple-layer UV images to store the Gaussian attributes during 3DGS fitting.
To constrain the 3D Gaussians located on the discrete rays, we propose a UV-based Adaptive Density Control.
We also use a stream-based training schema based on keyframes (image with yellow border).
% \srinath{Say the differences between the 3 ways and why it matters.}
% \srinath{Also, put the representation on top, fitting methd on the bottom to reflect the method section. Make sure to change refrences to top/bottom in the text}
}
% \vspace{-0.4cm}
\label{fig:method}
\end{figure*}

% \vspace{-2mm}
\section{Preliminaries}
\label{sec:Preliminaries}

% \subsection{Preliminaries}
% \vspace{-2mm}
We first provide a brief background on 3D Gaussian Splatting~\cite{kerbl3Dgaussians} and UVGS~\cite{rai2025uvgs} as our method builds upon them.

% \vspace{-4mm}
\paragraph{3D Gaussian Splatting ((3DGS))~\cite{kerbl3Dgaussians}:}
% \srinath{The para below is sloppy and does not make much sense. It is not describing 3dGS, just the Gaussian function.}
%
3DGS is a representation of 3D shape and appearance consisting of a set of Gaussian primitives with
%
% \begin{equation}
%     g(x) = \exp\left(-\frac{1}{2} (x - \mu)^\top \Sigma^{-1} (x - \mu)\right),
% \end{equation}
%
a position \( \mu \in \mathbb{R}^3 \), and covariance \( \Sigma \).
% , \srinath{what about color?}.
% To guarantee positive semi-definiteness during optimization, the covariance \( \Sigma \) is decomposed into a rotation matrix \(R \in  \mathbb{R}^{3\times3}\) and a scaling matrix \(S \in  \mathbb{R}^{3\times3}\):
% \(
    % \Sigma = R S S^\top R^\top.
% \)
% In practice, the rotation and scale matrices are represented by a unit quaternion \(r\in \mathbb{R}^4\) and diagonal scale vector \(s\in \mathbb{R}^3\), respectively.
Additionally, each 3D Gaussian primitive explicitly encodes view-dependent appearance via spherical harmonics (SH) coefficients \(c\) and an opacity value \(o\in \mathbb{R}\).
These attributes are optimized by minimizing the loss between the rendered and reference images.
To render a viewpoint, the Gaussians are projected as 2D splats and combined with $\alpha$-blending using a tile-based rasterizer. 
% \srinath{Say 1 or 2 sentences about how the primitives are optimized.}

\vspace{-4mm}
\paragraph{UV-based Point-to-Image Transformation.}
% \tao{the preliminary of UV-GS is too long. re-orgnize it by: (1) the core process of UV mapping; (2) it suceeds in object level reconstruction; (3) failed to model scene level task.}
The original 3DGS representation consists of an unstructured set of permutation-invariant primitives.
% \srinath{What is structural correspondence? Why not just say, it's `unstructured'.}
This unstructured nature poses challenges for downstream tasks, particularly when dealing with thousands or millions of Gaussians.

UVGS~\cite{rai2025uvgs} addresses this by introducing a structured UV-based point-to-image transformation that reformulates 3D Gaussians from an unordered set into a spatially organized UV image via spherical projection.
Each Gaussian \( g_i \) centered at \( \mu_i = (x_i, y_i, z_i) \) is transformed into spherical coordinates \( (\rho_i, \theta_i, \phi_i) \), where 

\(\rho_i = \sqrt{x_i^2 + y_i^2 + z_i^2}\), 
\( \theta_i = \tan^{-1}(y_i, x_i) \), 
and 

\( \phi_i = \cos^{-1}\left(\frac{z_i}{\rho_i}\right) \).
The azimuthal angle \( \theta_i \) and polar angle \( \phi_i \) are normalized to discrete UV coordinates in a map of size \( M \times N \):
% \vspace{-2mm}
\begin{equation}
\label{eq:uvgs2}
    u_i = \left\lfloor \frac{\pi + \theta_i}{2\pi} \times M \right\rfloor, \quad 
    v_i = \left\lfloor \frac{\phi_i}{\pi} \times N \right\rfloor.
% \vspace{-2mm}
\end{equation}
% \srinath{Is there a way to shorten the above to keep only essential info?}
Since multiple Gaussians may project to the same UV coordinate, UVGS uses multiple layers (\(K\) layers) to store top primitives, ordering them by opacity \(o\) within each pixel.
The final UV mapping is \( f: \text{UVCoords}\times \text{LayerIdx} \to \mathbb{R}^{D} \), where \(D\) encompasses all Gaussian attributes:
\(
    f(u, v, k) = \{\rho,~r,~s,~o,~c\} \in \mathbb{R}^{D} .
\)

The transformed UVGS representation introduces spatial coherence and resolves the permutation invariance problem.
% , as any arbitrary arrangement of 3DGS points now maps to the same UVGS representation. 
Interestingly, due to opacity-based sorting during the Gaussian mapping process, UVGS can effectively recover the surface-level Gaussians. 
\emph{The capability to represent a set of 3D Gaussian primitives in 2D while preserving surface-level details is of particular importance to our work.} 

However, it should be noted that this post-optimization UVGS mapping is highly lossy, since it projects only the Gaussian centers (mean positions) onto the UV space. 
As a result, it performs well for simple 3D objects but applying this post-hoc transformation to pretrained 4D Gaussian sequences often degrades visual quality, causing missing details and temporal inconsistencies (see supplementary).
% \srinath{Are we showing any experiments on the above claim?}
% The core issue is that 3DGS fitting for complex scenes produces Gaussian distributions that are not perfectly aligned with scene geometry, leading to substantial information loss during post-transformation pruning.
% \srinath{last few sentences are great. split into separate para and make it more concise.}

% \vspace{-2mm}
\section{Method}
%
% \srinath{Add a small para clearly outlining the goals in the method design choices. Then list mapping of those to subsections below.}
%
% \vspace{-2mm}
Our goal is to retain the structural benefits of UV mapping, preserve 3DGS's strong reconstruction quality, and capture 4D dynamic scenes.
To achieve this, we first propose \methodname-- a novel representation that combines a UV mapping strategy with progressive downscaling to represent 4D volumetric video.
% direct UV-constrained optimization with progressive storage strategies and compatibility with the existing multimedia infrastructure.
Second, we propose a novel fitting method, \fittername~(\Cref{sec:packuv-gs}) that uses optical flow based keyframing and Gaussian labeling to ensure smooth and lossless temporally-consistent reconstruction over long-horizon videos. 
% \srinath{what about temporal consistency?}
Our approach ensures efficient representation, accurate reconstruction, and scalability to arbitrary length in complex dynamic environments.

% \tao{move sec 3.3 ahead of how to fit and refine UV map, because the process of fitting uv map is too detailed.}

% \vspace{-2mm}
\subsection{\methodname Atlas}
\label{subsec:framepack_atlas}

% \srinath{I would not use the term Framepack. We are not packing frames, we are packing UV layers.}

% \srinath{It's very confusing to talk about the direct fitting first. I would simply descrieb the representation and motivation. Then in 4.2 talk about direct fitting.}
% \rai{I was thinking of this as a continuation of UVGS. In the previous section, we explained why post-optimization in UVGS fails, so this is a natural extension of that. Does that make sense or shall I movie it to 4.2?}
% \srinath{Without first describing the PackUV represenattion, fitting makes no sense. I would move it to 4.2.}

\paragraph{Pyramid UV Mapping:}
While direct UV optimization produces structured Gaussian maps, storing all \(K\) layers at uniform resolution \(M \times N\) incurs significant memory overhead—particularly problematic for high-resolution dynamic sequences.
However, we made an important observation: after sorting by opacity, \textbf{deeper layers (higher \(K\)) contain progressively fewer visible Gaussians} due to occlusion and opacity-based sorting across datasets (more details in the supplementary).
This means that not all layers are equally important, and we can therefore adopt a \emph{progressive, pyramid-like representation} (see \Cref{fig:method}, top). 
% combined with efficient \emph{atlas packing}.
Instead of storing all \(K\) layers at base resolution \(M_0 \times N_0\), we apply geometric downsampling that alternates dimension reduction:
\vspace{-3mm}
\begin{equation}
(M_k, N_k) = \begin{cases}
(M_0, N_0), & k = 0 \\
(M_{k-1}, N_{k-1}/2), & k \text{ odd} \\
(M_{k-1}/2, N_{k-1}), & k \text{ even}\nonumber
\end{cases}.
\vspace{-2mm}
\end{equation}
% \vspace{-1mm}
%
As shown in \Cref{fig:method}~(b), this pattern, \{$M_0 \times N_0$, $M_0 \times N_0/2$, $M_0/2 \times N_0/2$, $M_0/2 \times N_0/4$, \ldots\}, reflects increasing sparsity in deeper layers.
% Downsampling uses mode-based pooling to preserve dominant point indices, avoiding duplicate splats.

\vspace{-4mm}
\paragraph{UV Atlas Layout}
To maximize compactness, we pack the \(K\) progressive layers into a single texture atlas \(\mathcal{A}\) via recursive subdivision in layout that resembles a quadtree~\cite{framepack} (see \Cref{fig:method}~(c)):
\begin{itemize}[leftmargin=*,noitemsep,topsep=2pt]
\item Layer 0: Occupies right region at full resolution \(M_0 \times N_0\).
\item Layers 1--2: \(L_1\) (rotated \(90^\circ\) CCW) and \(L_2\) (horizontal) subdivide the left region. 
% \srinath{what resolution?}
\item Layers 3+: Continue recursive packing rightward of \(L_2\), alternating orientation (odd layers rotated, even horizontal) at progressively finer resolutions.
\end{itemize}
This packing technique achieves \textbf{88.5\% efficiency} (utilized pixels / total atlas pixels) while significantly outperforming grid (\(\sim\)60\%) or pyramid (\(\sim\)75\%) layouts.
The atlas dimensions we use are:
%
% \vspace{-5mm}
\begin{equation}
W_\mathcal{A} = N_0 + \sum_{k=1}^{K-1} N_k, \quad H_\mathcal{A} = \max_{k} M_k.\nonumber
% \vspace{-3mm}
\end{equation}

Finally, to represent volumetric video, we assemble a continuous sequence of such UV atlases, each one representing 1 video frame.
\methodname is seamlessly integrated into our training pipeline: during optimization, UV maps are maintained at their respective progressive resolutions, and upon convergence, layers are packed into the atlas for efficient storage and streaming (Section~\ref{sec:packuv-gs}).

% \begin{figure}[th]
%     \centering
%     \includegraphics[width=0.9\linewidth]{images/fig_pruning.pdf}
%     \vspace{-0.4cm}
%     \caption{UV-based Pruning. During the location update of the 3D Gaussian \textcolor{orange}{\huge \textbullet} in training, we enforce that it always allocates on one of the discrete rays corresponding to each UV pixel.
%     After updating, the 3d Gaussian will (a) still map to the original UV pixel, or (b) be pruned since no corresponding UV pixel, or (c) map to another pixel of the original UV pixel, according to \cref{eq:uvgs1,eq:uvgs2}.
%     The green areas indicate the near region with $\delta(\rho_i)$ threshold.}
%     \label{fig:uv_pruning}
%     \vspace{-0.4cm}
% \end{figure}

% \subsection{Temporal Fitting Pipeline}
\subsection{\fittername Fitting}
\label{sec:packuv-gs}
% \srinath{Temporal Fitting Pipeline --> \fittername}

% \srinath{One sentence first about the goal of \fittername} \rai{Todo}
% \srinath{The direct fitting method should go here}

The goal here is to efficiently fit PackUV directly from multiview videos using optical-flow-based keyframing and Gaussian labeling to handle large motions, disocclusions, and maintain temporal consistency. 

\vspace{-4mm}
\paragraph{Fitting UV Maps Directly.}
\label{sec:fittingUV}
Instead of first fitting 3DGS and then projecting to the UV space~\cite{rai2025uvgs}, we optimize the scene Gaussians directly within UV space with fixed spatial resolution and predetermined layer count (K).
Let \( U \in \mathbb{R}^{M \times N \times K \times D} \) denote the UV maps, where \(M \times N\) defines the UV grid resolution, \(K\) is the number of layers per pixel, and \(D\) encodes all Gaussian attributes:
% \vspace{-3mm}
\begin{equation}
U[u_i, v_i, k] = g_i = \{\rho_i,~r_i,~s_i,~o_i,~c_i\} \in \mathbb{R}^{D}.
% \vspace{-2mm}
\end{equation}

This direct optimization not only preserves the structural benefits for downstream tasks but also enforces Gaussian sparsity through the discrete UV grid structure.

\subsubsection{Video Keyframing}
To efficiently fit given synchronized multi-view videos with $T$ frames, represented as a set $\{V_n\}_{n=1}^N$, where N is the total number of views.
We divide each video into s set of $m$ temporal segments.
To do so, for each frame $t$, we compute the optical-flow magnitude $M(t)$ on one video, select the top $(m-1)$ magnitude peaks with a minimum separation $\theta$, and use the first frame of every segment as a keyframe. 
% \srinath{Shoudn't keyframing be done before diving into temporal segments?}
% \[F^K_0 = 1, \qquad\] 
% \[ \{F^K_i\}_{i=1}^{m-1} = \operatorname{Top}_{m-1}\big(M(t)\big) \ \text{s.t.}\ |F^K_i - F^K_{i-1}| \ge \tau . \]
% \srinath{Notation above is too cumbersome, please simplify.}
These keyframes define the segment boundaries. 
For each keyframe $F^K_i$, the PackUV Gaussians are initialized from the previous keyframe which preserves temporal and spatial consistency.
The frames between keyframes are treated as transition frames. 
Each transition frame $F^t$ is initialized from the preceding frame and refined with a few training iterations compared to $F^K$:
% \vspace{-3mm}
\[
\mathcal{G}(K) \leftarrow \text{Update}\!\left(\mathcal{G}(K-1)\right),
\mathcal{G}(t) \leftarrow \text{Update}\!\left(\mathcal{G}(t-1)\right).
\]

% \vspace{-2mm}
This staged, stream-based strategy enables efficient reconstruction of high-fidelity dynamic scenes while allowing us to parallelize the fitting process.
Frames exhibiting high drift, occlusions/disocclusions, or appearance breaks are promoted to keyframes.
This keyframing technique helps us handle arbitrary length sequences, large motions, and disocclusions without quality degradation over time.
More details are given in the supplementary.
\subsubsection{Gaussian Labeling}
\label{sec:gaussian_labeling}

On top of sequential keyframing pipeline for dynamic Gaussian splatting, we also use optical flow to isolate dynamic regions and freezes static Gaussians during optimization. Flow is computed per camera via RAFT~\cite{raft_flow}, and a CUDA-accelerated covariance-aware projection robustly determines which 3D Gaussians overlap with flow-detected dynamics. The approach improves temporal stability and training efficiency on multi-sequence videos while preserving static backgrounds.

\begin{figure*}[!h]
\centering
\includegraphics[width=6.8in]{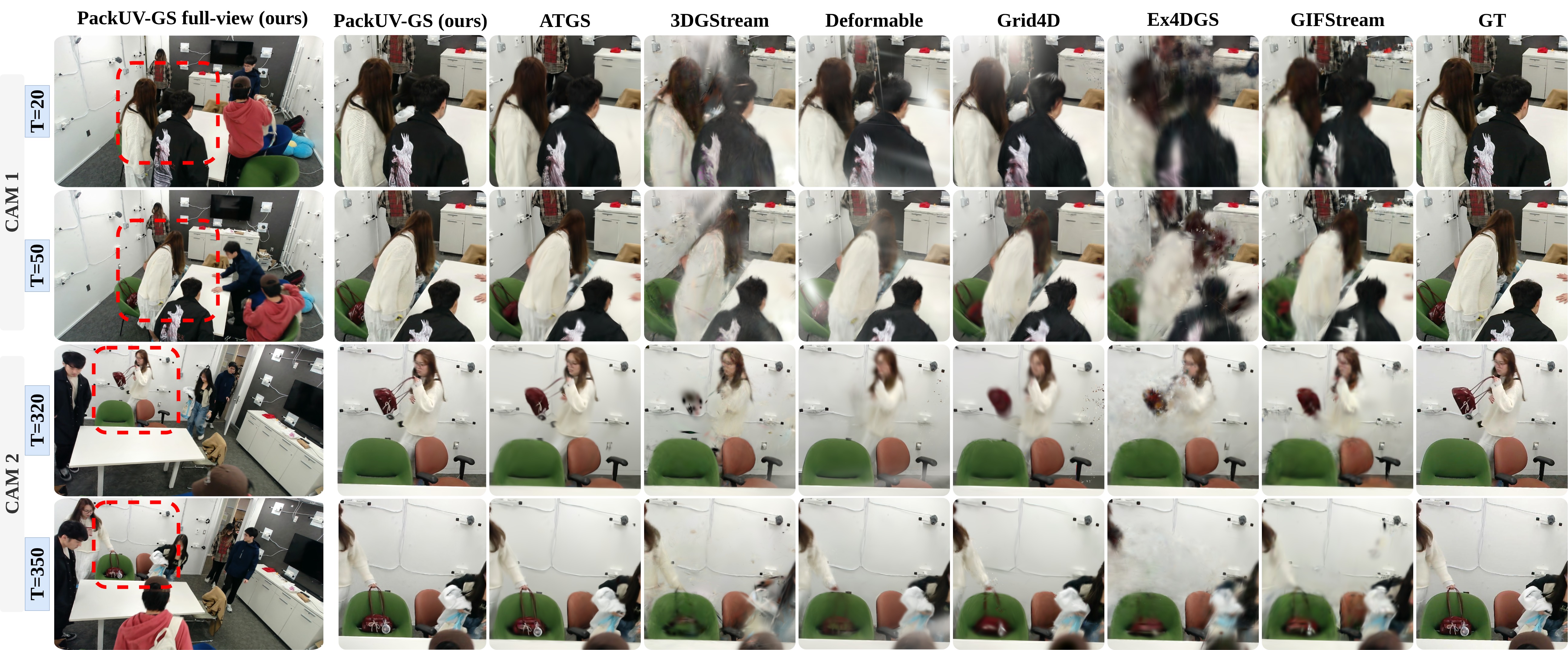} 
\vspace{-0.2cm}
\caption{
\fittername vs. baselines for large motion and disocclusion handling. The proposed keyframing and Gaussian labeling strategy effectively manages complex scenarios, such as new objects or people entering a room and dispersing. Zoom to view better.
\vspace{-0.2cm}
}
\label{fig:disocclusion}
\end{figure*}

\vspace{-4mm}
\paragraph{Optical Flow and Binary Motion Masks.}
\label{subsec:raft-flow}
For each camera view $c$, we estimate forward optical flow $\mathbf{F}^{c}_{(t-1)\rightarrow t}$ between consecutive frames $(I^{c}_{t-1}, I^{c}_{t})$ using RAFT~\cite{raft_flow}.
% %
% \begin{equation}
%   \mathbf{F}^{c}_{t-1\rightarrow t} \;=\; \mathrm{RAFT}\bigl(I^{c}_{t-1}, I^{c}_{t}\bigr) \in \mathbb{R}^{H\times W\times 2}.\nonumber
% \end{equation}
%
We form a binary motion mask by thresholding the flow magnitude and dilating to include local context:
% \begin{equation*}
%   M^{c}_{t}(\mathbf{p}) \,=\, \mathbb{1}\Bigl(\bigl\lVert \mathbf{F}^{c}_{t-1\rightarrow t}(\mathbf{p}) \bigr\rVert_2 \,>\, \tau\Bigr)\,,
% \end{equation*}
\vspace{-3mm}
\[
M_t^c(\mathbf{p}) = 
\begin{cases}
1, & \|\mathbf{F}^c_{t-1 \rightarrow t}(\mathbf{p})\|_2 > \tau,\\[4pt]
0, & \text{otherwise}.
\end{cases}
\vspace{-2mm}
\]
% \vspace{-1mm}
\[
  \qquad M^{c}_{t} \leftarrow \mathrm{dilate}\bigl(M^{c}_{t};\, r\bigr)\,.
\]
% \srinath{Notation again. I have no clue what that symbol above is.}
$\tau$ is the flow magnitude threshold and $r$ the dilation radius.
% We compute and cache $M^{c}_{t}$ per camera for the whole segment, and masks are reused across epochs \srinath{what is an epoch here?}.

\vspace{-4mm}
\paragraph{Covariance-Aware Gaussian Masking:}
\label{subsec:masking}
Each Gaussian $g_i$ has mean $\boldsymbol{\mu}_i\in\mathbb{R}^3$, diagonal scales $\mathbf{s}_i\in\mathbb{R}^3$, and rotation $\mathbf{q}_i$ (unit quaternion). Its 3D covariance is represented using
% \[
%   \Sigma^{3D}_i \;=\; \mathbf{R}(\mathbf{q}_i)\; \mathbf{S}(\mathbf{s}_i)\, \mathbf{S}(\mathbf{s}_i)^\top\; \mathbf{R}(\mathbf{q}_i)^\top\,,
% \]
$\mathbf{R}(\cdot)$ is the rotation matrix and $\mathbf{S}(\mathbf{s})=\mathrm{diag}(\mathbf{s})$ \cite{kerbl3Dgaussians}.

Let $\mathbf{T}_c$ denote the camera $c$'s view transformation matrix and $\mathbf{J}_c(\cdot)$ its $2\times 3$ Jacobian evaluated at the camera-space mean. Following EWA splatting~\cite{ewasplatting}, the 2D covariance of $g_i$ in image space is
% \vspace{-2mm}
\begin{equation}
  \Sigma^{2D}_{i,c} \;=\; \mathbf{J}_c\, \Sigma^{3D}_{i,\mathrm{cam}}\, \mathbf{J}_c^\top\,,
  \qquad \Sigma^{3D}_{i,\mathrm{cam}} = \mathbf{T}_c\, \Sigma^{3D}_i\, \mathbf{T}_c^\top\,.
\end{equation}

% \srinath{I don't think the Gaussian parameter description above is needed. Just point to 3DGS paper. Only mention the projected Gaussian.}
% \vspace{-2mm}
We then project the mean to normalized device coordinates (NDC), obtain pixel coordinates $\mathbf{m}_{i,c} \in \mathbb{R}^2$, and test overlap with the motion mask using the Mahalanobis metric~\cite{mahalanobis1930tests}. A pixel $\mathbf{p}$ is inside the ellipse if
% \vspace{-3mm}
\[
  d^2\!(\mathbf{p};\, \mathbf{m}_{i,c},\, \Sigma^{2D}_{i,c}) \,=\, (\mathbf{p}-\mathbf{m}_{i,c})^\top \bigl(\Sigma^{2D}_{i,c}\bigr)^{-1} (\mathbf{p}-\mathbf{m}_{i,c}) \;\le\; 9\,.
\]
A Gaussian is marked dynamic for camera $c$ if any pixel within a radius derived from the ellipse’s largest eigenvalue satisfies $M^{c}_{t}(\mathbf{p})=1$:
% \vspace{-2mm}
\begin{equation}
  D_{i,c} \;=\; \bigvee_{\mathbf{p}\in \mathcal{E}_{i,c}} M^{c}_{t}(\mathbf{p})\,, \quad \mathcal{E}_{i,c} = \{\mathbf{p}\,|\, d^2\!(\mathbf{p}) \le 9\}\,.
% \vspace{-3mm}
\end{equation}
% \vspace{-4mm}
The final dynamic mask across cameras is an OR aggregation:
% \vspace{-4mm}
\[
  D_i \;=\; \bigvee_{c\in\mathcal{C}} D_{i,c}\,.
  % \vspace{-2mm}
\]
To make the computations realtime, we implement the above in a custom CUDA kernel that, for each Gaussian, 
(i) computes $\Sigma^{2D}_{i,c}$ via the projection Jacobian, 
(ii) estimates a sampling radius from the largest eigenvalue, and (iii) scans pixels within this radius, testing $d^2\!\le 9$ and $M^{c}_{t}(\mathbf{p})=1$. 

% If CUDA is unavailable, we can fall back to a robust point-projection heuristic $D_{i,c}=M^{c}_{t}(\Pi_c(\boldsymbol{\mu}_i))$.

% \vspace{-4mm}
\paragraph{Gradient Freezing.} During backpropagation we zero gradients of all static Gaussians, i.e.,

\(
  \nabla_{\boldsymbol{\theta}_i} \mathcal{L} \;\leftarrow\; D_i\, \nabla_{\boldsymbol{\theta}_i} \mathcal{L}\,, \qquad D_i\in\{0,1\},
\)

\noindent
where $\boldsymbol{\theta}_i$ includes position, scale, rotation, color, and opacity. We additionally reset optimizer momentum for static Gaussians periodically to prevent drift. When densifying, child Gaussians inherit the parent's dynamic/static label to preserve the dynamic ratio across training.

% \srinath{Shouldn't this pruning go to the fitting part?}
\subsubsection{UV-Based Pruning}
To reduce redundancy and keep discretized UV mapping during optimization, we introduce two pruning strategies: 
% (illustrated in Figure~\ref{fig:uv_pruning}):

\begin{itemize}[leftmargin=*,topsep=3pt]
    \item \textbf{Valid UV Projection Pruning.}  
    After densification, we recalculate each child Gaussian's UV coordinates \((u_{\text{scaled}}, v_{\text{scaled}})\).
    Gaussians failing to satisfy equation~\ref{eq:uvgs2} are pruned.
    This structural pruning enforces sparsity while aligning Gaussians more closely with scene geometry, improving both efficiency and visual quality.

    \item \textbf{Max-\(K\) UV Pruning.}  
    To prevent overpopulation at individual UV pixel, we retain only the top \(K\) Gaussians per pixel based on opacity.
    Formally, for each coordinate \((u,v)\) with associated Gaussians \(\mathcal{G}_{uv}\):
    
    \(
        |\mathcal{G}_{uv}| > K \quad \Rightarrow \quad \mathcal{G}_{uv} \leftarrow \text{TopK}(\mathcal{G}_{uv}, K)
    \)
\end{itemize}

These pruning techniques reduce memory overhead, focus learning on surface-relevant regions, and improve reconstruction quality while accelerating optimization.

% \srinath{Remove equation numbers unless you re-use them somewhere. Use nonumber}

\subsubsection{Objective}
%
% \srinath{This need not be a section, sub-section under \fittername is fine.}
Let $\hat{I}^{c}_t$ be the rendered image for camera $c$ and ground-truth $I^{c}_t$. The photometric objective blends L1 and SSIM:
% \vspace{-2mm}
\begin{equation}
% \vspace{-2mm}
  \mathcal{L}_{\mathrm{photo}} \;=\; (1-\lambda_{\mathrm{ssim}})\, \lVert \hat{I}^{c}_t - I^{c}_t \rVert_1 
  \; +\; \lambda_{\mathrm{ssim}}\, \bigl(1-\mathrm{SSIM}(\hat{I}^{c}_t, I^{c}_t)\bigr)\,.
  \nonumber
\end{equation}

To discourage floaters and oversized primitives, we regularize scales and opacities, optionally restricted to dynamic Gaussians ($D_i{=}1$):
% \vspace{-4mm}
\begin{align}
  \mathcal{L}_{\mathrm{scale}} &\;=\; \mathbb{E}_i\, \Bigl[\max\{0,\; \max(\mathbf{s}_i) - s_\mathrm{max}\}\Bigr]^2\,,\\
  \mathcal{L}_{\mathrm{opacity}} &\;=\; \mathbb{E}_i\, \alpha_i\, (1-\alpha_i)\,.
  % \vspace{-2mm}
\end{align}
% \vspace{-3mm}
% \[
% \mathcal{L}_{\text{scale}} = 
% \Bigl[\mathbb{E}_i\, \max\bigl(0,\; \max(\mathbf{s}_i) - s_{\max}\bigr)\Bigr]^2,
% \]
% \[
% \mathcal{L}_{\text{opacity}} = 
% \mathbb{E}_i\, \alpha_i(1-\alpha_i).
% \]
The total loss is
\begin{equation}
  \mathcal{L} \;=\; \mathcal{L}_{\mathrm{photo}} \; + \; \mathcal{L}_{\mathrm{depth}} \; + \; \lambda_{\mathrm{scale}}\, \mathcal{L}_{\mathrm{scale}} \; + \; \lambda_{\mathrm{opacity}}\, \mathcal{L}_{\mathrm{opacity}}\,.
  \nonumber
  % \vspace{-2mm}
\end{equation}

\vspace{-4mm}
\paragraph{Low Precision Optimization.}
%
% Unlike most prior works that employ lossy quantization after optimization for storage efficiency, \fittername utilizes low precision optimization (LPO) for learning the Gaussian attributes 
% $\theta = \{\mathbf{\mu}, \mathbf{s}, \mathbf{r}, \alpha, \mathbf{c}\}$.
% Our experiments show that LPO effectively compensates for the quantization loss, maintaining both PSNR and training efficiency.
% At each iteration, the renderer operates on a quantized proxy $\tilde{\theta}$ obtained by a uniform $K$-bit fake quantizer with scale $\Delta$.
% We use a straight-through estimator (STE) to preserve gradient flow while keeping master weights in FP32. 
% The training objective remains the same, combining an L1+SSIM photometric term with a scheduled regularizer.
% Backpropagation updates FP32 master parameters with Sparse Adam optimizer~\cite{taming3dgs}.

% We use 8-bit reduced precision for $\{\mathbf{s}, \mathbf{r}, \alpha, \mathbf{c}\}$ and 16-bit for $\{\mathbf{x}\}$. 
% During storage, the 16-bit $\{\mathbf{x}\}$ values are split into two 8-bit parts. 
% This 8-bit-per-channel design makes \methodname readily compatible existing video coding infrastructures, enabling the direct application of both lossless or lossy compression methods (\eg,~HEVC, AVC, FFV1).

Unlike prior methods that quantize Gaussian parameters only after training, \fittername performs low-precision optimization (LPO) directly over the attributes $\theta={\mathbf{\mu},\mathbf{s},\mathbf{r},\alpha,\mathbf{c}}$.
At each iteration, the renderer consumes a uniformly quantized $K$-bit proxy $\tilde{\theta}$, while gradients flow through a straight-through estimator and FP32 master weights. The training loss (L1+SSIM with a scheduled regularizer) and optimizer (Sparse Adam~\cite{taming3dgs}) remain unchanged.
LPO compensates quantization error, preserving both PSNR and training speed. We use 8-bit precision for ${\mathbf{s},\mathbf{r},\alpha,\mathbf{c}}$ and 16-bit for ${\mathbf{x}}$, later split into two 8-bit channels for storage. This 8-bit–per-channel layout is compatible with standard video codecs (e.g., HEVC, AVC, FFV1), enabling both lossless and lossy compression.

\newcolumntype{L}[1]{>{\raggedright\arraybackslash}p{#1}}
\newcolumntype{C}[1]{>{\centering\arraybackslash}p{#1}}
\newcolumntype{R}[1]{>{\raggedleft\arraybackslash}p{#1}}
\begin{table*}[!h] % h: here, t: top, b: bottom, p: page of floats
  \centering 
  \fontsize{8pt}{10pt}\selectfont
  \vspace{-0.3cm}
  \caption{\textbf{Dataset Comparisons.}
We compare our newly captured dataset \datasetname with existing multi-view datasets across sequence count, total frames, camera setup, resolution, maximum FPS, scenario type, and view range.
\datasetname contains \numsequences diverse sequences totaling over \numimgs~(billion) high-quality frames, recorded with more than 50 cameras at $1920 \times 1200$ resolution.
The capture system supports up to 90 FPS, providing high temporal fidelity.
% Finally, 'Scenario Type' and 'View Range' respectively define the category of recorded scenes and the extent of scene coverage.
  }
  % \vspace{-0.3cm}
  \resizebox{0.95\linewidth}{!}{
  \begin{tabular}{@{}L{4cm}C{1cm}C{1.2cm}C{1.5cm}C{2cm}C{1cm}C{2cm}C{2cm}@{}}
\toprule
Dataset & Sequence & Frames & Camera & Resolution & \begin{tabular}[c]{@{}c@{}}Max FPS\end{tabular} & Scenario & \begin{tabular}[c]{@{}c@{}}View Range\end{tabular} \\
\midrule

D-NeRF~\cite{pumarola2020d}
& \cellcolor{white!40}8
& --
& --
& $800 \times 800$
& --
& Simulator
& \cellcolor{white!40}360 \\

CMU Panoptic~\cite{Joo_2015_ICCV}
& \cellcolor{white!40}65
& \cellcolor{white!40}$\sim 15$M
& \cellcolor{white!40}31 (+ 480$^*$)
& $1920 \times 1080$
& 30
& Real-World
& \cellcolor{white!40}360 \\

N3DV~\cite{li2022neural}
& 6
& $\sim 38$K
& 21
& \cellcolor{white!40}$2028 \times 2704$
& 30
& Real-World
& face forward \\

NeRF-DS~\cite{yan2023nerf}
& \cellcolor{white!40}8
& $\sim 8$K
& 2
& $480 \times 270$
& --
& Real-World
& face forward \\

SelfCap~\cite{xu2024representing}
& 3
& \cellcolor{white!40}$\sim 1.5$M
& \cellcolor{white!40}22
& \cellcolor{white!40}$3840 \times 2160$
& \cellcolor{white!40}60
& Real-World
& face forward \\

\midrule

\datasetname
& \cellcolor{white!40}\textbf{100}
& \cellcolor{white!40}\textbf{$\sim$ 2B+}
& \cellcolor{white!40}\textbf{55--88}
& \cellcolor{white!40}$1920 \times 1200$
& \cellcolor{white!40}\textbf{90}
& Real-World
& \cellcolor{white!40}\textbf{360} \\

\bottomrule
\end{tabular}
    }
  \label{tab:dataset_comparison} 

  \vspace{-0.2cm}
\end{table*}

% \vspace{-2mm}
\section{\datasetname Dataset}
\label{sec:dataset}

Existing multi-view datasets~\cite{Joo_2015_ICCV,li2022neural,yan2023nerf,xu2024representing,pumarola2020d} mostly use front-facing cameras and offer limited diversity, making them insufficient for evaluating volumetric video methods under fast motion, large deformation, and disocclusion.
We introduce \datasetname, a real-world, long-horizon multi-view dataset comprising \numsequences dynamic sequences and over \numimgs frames, captured with 55-88 synchronized cameras. The data spans studio and in-the-wild settings, including human–human, human–object, and robot–object interactions. 
Sequences average 10 minutes and reach up to 30 minutes.
To form a comprehensive benchmark, \datasetname includes activities with broad variation in motion speed (from slow pouring water to fast basketball and dance), motion scale (from table-top manipulation to pickleball), and object properties (rigid, articulated, reflective, transparent).
To our knowledge, \datasetname surpasses existing datasets in sequence length, camera count, motion complexity, dynamic diversity, and data volume.
We present detailed comparisons between \datasetname and existing datasets in \Cref{tab:dataset_comparison}.
% The advantages in Sequence, Frames, Camera, and View Range indicate that \datasetname provides substantial improvements and extensions across both spatial and temporal dimensions.
We plan to release \datasetname publicly, with the aim of establishing it as a new benchmark for evaluating general-purpose, long-horizon dynamic reconstruction.
Additional details about \datasetname can be found in the supplementary material.

% \vspace{-0.3cm}

% \vspace{-3mm}
\section{Experiments}
\label{sec:experiments}

\begin{figure*}[!ht]
\centering
\vspace{-2mm}
\includegraphics[decodearray={-0.2 1.2 -0.2 1.2 -0.2 1.2},width=6.8in]{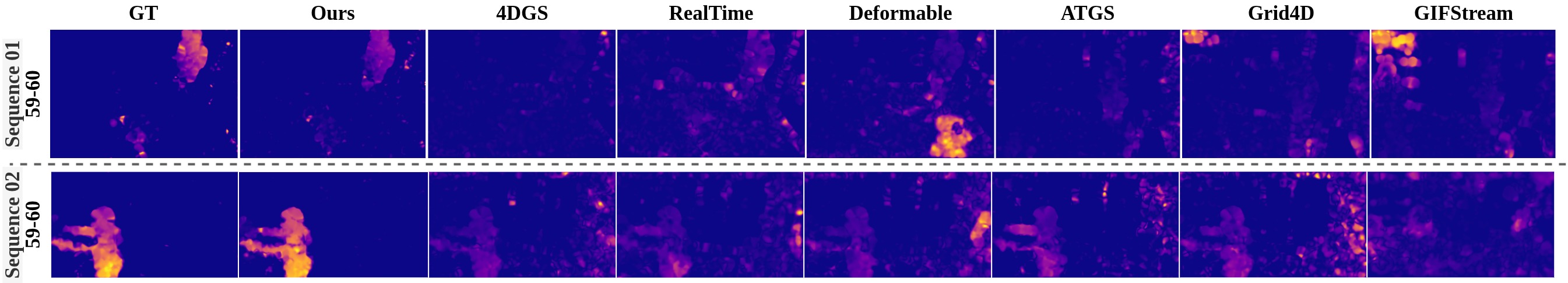} 
\vspace{-2mm}
\caption{
Optical flow. To assess long-term temporal stability, we compute optical flow between consecutive timestamps. 
% \tao{add the rgb image of frame 59 and 60?}
}
\vspace{-2mm}
\label{fig:optical_flow}
\end{figure*}

% \vspace{-2mm}
% \subsection{Setup}
% \paragraph{Setup.}

In this section, we conduct comprehensive experiments on three different datasets to verify the effectiveness of our method in generating long volumetric videos with large motion and disocclusion while being compatible with the existing video coding infrastructure.
Section~\ref{sec:qual_quat} discusses the qualitative and quantitative evaluations of our proposed method and compare it to the baselines.
In Section~\ref{sec:video_coding}, we show how \methodname can be losslessly streamed via existing video infrastructure.
The ablation study in Section~\ref{sec:ablation} shows the effectiveness of each component in our method.

\vspace{-5mm}
\paragraph{Setup.} We set the \methodname atlas resolution $M_0$ and $N_0$ to 1024, and the number of UV layers $K$ to 8. $\theta$ for keyframing is set to 30. 
% $Densify\_Grad\_Threshold$ was set to 0.0005 for $F^K$ and 0.02 for $F^t$. 
$\lambda_\mathrm{scale}$ and $\lambda_\mathrm{opacity}$ were set to 0.0001.

\vspace{-4mm}
\paragraph{Dataset.}
We conducted experiments on widely used real-world datasets, N3DV~\cite{li2022neural} and SelfCap~\cite{xu2024representing}, as well as our \datasetname.
The N3DV dataset is collected from 21 cameras to capture the central scene from the face-forward views at 2704$ \times$2028 resolution and 30~FPS.
% The videos are captured at 2704$ \times$2028 resolution and 30~FPS. 
We select the \textit{flame\_steak} sequence for evaluation and set aside a testing camera following~\cite{li2022neural}.
The SelfCap dataset contains longer sequences, captured with 22 cameras at 4K resolution and 60 FPS. 
We evaluate on two sequences, \textit{hair\_release} and \textit{yoga}, and reserve 2 cameras for testing. 
We additionally select 5 sequences from \datasetname containing multiple subjects and complex motion: \textit{baby\_dance}, \textit{spot}, \textit{entering\_room}, \textit{object\_place}, \textit{kitchen}, and \textit{entering\_room}. 
For all datasets, we downsample to 1.6K resolution following~\cite{kerbl3Dgaussians}. 

\vspace{-4mm}
\paragraph{Baselines.}
We compare with several state-of-the-art baseline methods, 3DGStream~\cite{sun20243dgstream}, 4DGS~\cite{Wu_2024_CVPR}, RealTime4DGS~\cite{yang2023real}, Deformable3DGS~\cite{yang2024deformable}, ATGS~\cite{chen2025adaptive}, Grid4D~\cite{xu2024grid4d}, Ex4DGS~\cite{lee2024fully}, and GIFStream~\cite{Li_2025_CVPR}.
% We compare with several state-of-the-art baseline methods
% \srinath{explicitly name the methods to make it easier for readers to udnerstand}
% \cite{sun20243dgstream},\cite{Wu_2024_CVPR}, \cite{yang2023real}, \cite{yang2024deformable}, \cite{chen2025adaptive}, \cite{xu2024grid4d}, \cite{lee2024fully}, and \cite{Li_2025_CVPR}.
We train 3DGStream on the full sequence length.
% In contrast, Deformable3DGS, RealTime4DGS, 4DGS, and Grid4D utilize deformation networks model dynamic scenes with 3D Gaussians, which leads to high VRAM consumption for longer sequences.
To mitigate high VRAM consumption in Deformable3DGS, RealTime4DGS, 4DGS, and Grid4D, we split the sequences into segments and train each segment independently.
% We employ a similar training technique for AT-GS. 
Although ATGS employs a streaming-framework, we observe that training on full sequences results in gradient explosion and adopt the segmented training strategy. 
We performed all the experiments on RTX 3090 GPU, 256 GB of RAM and 8 CPU cores. 
We evaluate rendering quality using average PSNR, SSIM, and LPIPS~\cite{zhang2018unreasonable} across all timestamps. 

\begin{table*}[!h]
\fontsize{8pt}{10pt}\selectfont
\centering
% \vspace{-2mm}
\caption{Quantitative Comparison. We report PSNR, SSIM, LPIPS, and train time (in hours) for a window length of 60 timestamps. We also report the method's streaming ability and compatibility with the existing video coding infrastructure.}
\vspace{-0.2cm}
\begin{adjustbox}{max width=\textwidth}
\resizebox{0.95\linewidth}{!}{
\begin{tabular}{
>{\centering\arraybackslash}p{1.6cm} |
>{\centering\arraybackslash}p{0.7cm} >{\centering\arraybackslash}p{0.7cm} >{\centering\arraybackslash}p{0.7cm} >{\centering\arraybackslash}p{0.7cm} |
>{\centering\arraybackslash}p{0.7cm} >{\centering\arraybackslash}p{0.7cm} >{\centering\arraybackslash}p{0.7cm} >{\centering\arraybackslash}p{0.7cm} |
>{\centering\arraybackslash}p{0.7cm} >{\centering\arraybackslash}p{0.7cm} >{\centering\arraybackslash}p{0.7cm} >{\centering\arraybackslash}p{0.7cm} |
>{\centering\arraybackslash}p{0.7cm} >{\centering\arraybackslash}p{0.7cm}
}
\toprule
\multirow{2}{*}{Method} &
\multicolumn{4}{c|}{\datasetname} &
\multicolumn{4}{c|}{SelfCap~\cite{xu2024longvolcap}} &
\multicolumn{4}{c|}{N3DV~\cite{li2022neural}} & \\
& PSNR$^\uparrow$ & SSIM$^\uparrow$ & LPIPS$^\downarrow$ & Train$^\downarrow$
& PSNR$^\uparrow$ & SSIM$^\uparrow$ & LPIPS$^\downarrow$ & Train$^\downarrow$
& PSNR$^\uparrow$ & SSIM$^\uparrow$ & LPIPS$^\downarrow$ & Train$^\downarrow$
& Stream & Codec\\
\midrule

3DGStream  
& \cellcolor{orange!40}23.17 & \cellcolor{orange!40}0.826 & \cellcolor{yellow!40}0.33 & \cellcolor{orange!40}1.00
& \cellcolor{yellow!40}19.77 & \cellcolor{orange!40}0.769 & \cellcolor{yellow!40}0.36 & \cellcolor{yellow!40}1.43
& 31.17 & 0.952 & \cellcolor{orange!40}0.23 & \cellcolor{red!40}0.31
& Full & No \\

4DGS 
& \cellcolor{yellow!40}23.11 & \cellcolor{yellow!40}0.808 & 0.35 & 2.30
& 19.56 & 0.745 & 0.37 & 3.18
& 29.81 & 0.951 & \cellcolor{yellow!40}0.21 & 3.27
& No & No \\

RealTime 
& 21.37 & 0.790 & 0.38 & 4.48
& 19.46 & \cellcolor{yellow!40}0.747 & 0.41 & 8.07
& \cellcolor{orange!40}32.29 & \cellcolor{red!40}0.955 & \cellcolor{yellow!40}0.21 & 2.48
& No & No \\

Deformable 
& 20.07 & 0.778 & \cellcolor{yellow!40}0.33 & 2.04
& 17.89 & 0.708 & 0.38 & 2.09
& 26.51 & 0.935 & 0.24 & 0.62
& No & No \\

ATGS 
& 21.42 & 0.796 & 0.36 & 1.13
& 15.48 & 0.664 & 0.51 & 1.82
& 30.99 & 0.934 & 0.24 & 1.97
& Pseudo & No \\

Grid4D 
& 21.58 & 0.790 & 0.37 & 1.13
& 17.53 & 0.701 & 0.44 & 1.82
& 30.87 & \cellcolor{yellow!40}0.954 & \cellcolor{red!40}0.197 & 1.97
& No & No \\

Ex4DGS 
& 20.73 & 0.789 & 0.39 & \cellcolor{red!40}0.83
& 17.62 & 0.680 & 0.39 & \cellcolor{orange!40}1.23
& \cellcolor{yellow!40}31.57 & 0.944 & \cellcolor{orange!40}0.23 & \cellcolor{yellow!40}0.59
& No & No \\

GIFStream 
& 21.92 & 0.795 & 0.39 & 1.61
& \cellcolor{orange!40}19.78 & 0.745 & \cellcolor{orange!40}0.35 & 2.05
& 31.10 & \cellcolor{yellow!40}0.954 & 0.25 & \cellcolor{orange!40}0.42
& Pseudo & Partial \\

Ours 
& \cellcolor{red!40}27.41 & \cellcolor{red!40}0.842 & \cellcolor{red!40}0.28 & \cellcolor{yellow!40}1.05
& \cellcolor{red!40}22.52 & \cellcolor{red!40}0.783 & \cellcolor{red!40}0.31 & \cellcolor{red!40}1.12
& \cellcolor{red!40}32.81 & \cellcolor{orange!40}0.953 & \cellcolor{yellow!40}0.21 & 1.37
& Full & Full \\

\bottomrule
\end{tabular}
}
\end{adjustbox}
\label{tab:qual_comp} 
\vspace{-0.2cm}
\end{table*}
% \cellcolor{orange!40}
% \cellcolor{yellow!40}
% \cellcolor{red!40}

% \vspace{-4mm}
% \paragraph{Metrics.}
% We evaluate rendering quality using average PSNR, SSIM, and LPIPS~\cite{zhang2018unreasonable} across all timestamps. 
% Additionally, we calculate the average training time for 60 timestamps (in hours), the rendering FPS, and the average storage for 60 timestamps (in GB). 
% For methods train in segments -- 4DGS, RealTime4DGS, Deformable 3DGS, and AT-GS -- we report the segment window length. 

% \begin{table}[t]
% \centering
% \setlength{\tabcolsep}{0.95mm}
% \renewcommand{\arraystretch}{1.2}
% \caption{We compressed the UV images in $8 \times 12 \times 1024^2$ first by the neural compressor[NC] (lossy), then zstd (lossless). 
% In zstd compression, we test two strategies: compress the residue and compress the original frame.}
% \vspace{-0.2cm}
% \resizebox{0.98\columnwidth}{!}{
% \begin{tabular}{l|ccccc}
%  \toprule
%  \textbf{UV Image Size} & \textbf{PSNR}$^\uparrow$ & \textbf{SSIM}$^\uparrow$ & \textbf{LPIPS}$^\downarrow$ & [NC]\textbf{Size}$^\downarrow$ & [+zstd]\textbf{Size}$^\downarrow$ \\
%  \midrule
%  $8 \times 1024^2$ & 26.01 & 0.85 & 0.29 & 33.6MB & 24.42MB/28.06MB\\
%  $3 \times 1024^2$ & 24.87 & 0.78 & 0.35 & 12.6MB & 9.1MB/10.52MB\\
%  $8 \times 512^2$ & 25.19 & 0.83 & 0.31 & 8.4MB & 6.1MB/6.9MB\\
%  $3 \times 512^2$ & 22.38 & 0.73 & 0.37 & 3.1MB & 2.32MB/2.59MB\\
%  \midrule
%  $8 \times 12 \times 1024^2$ & 27.14 & 0.87 & 0.27 & - & 53.2MB/62.9MB\\
%  \bottomrule
% \end{tabular}}
% \label{table:compress} 
% \vspace{-0.6cm}
% \end{table}

% \vspace{-2mm}
\subsection{Qualitative and Quantitative Results}
\label{sec:qual_quat}
\paragraph{Qualitative results.} \Cref{fig:disocclusion} shows qualitative comparisons where our method excels at handling large motions and severe disocclusions. 
Our keyframing and Gaussian-labeling strategy robustly manages complex events, including new objects or people entering and interacting (e.g., people entering a room and dispersing). 
Additional results appear in the supplementary material.
Deformable3DGS, RealTime4DGS, Grid4D, and 4DGS depend on deformation networks, leading to high VRAM usage, poor scalability to long sequences, and difficulty modeling newly emerging objects. 
% RealTime4DGS stores separate Gaussians per frame, incurring similarly heavy memory and storage overhead. 
Although streaming-based approaches can support longer sequences, they still produce from artifacts. 
Specifically, 3DGStream struggles with large motions, ATGS suffers from gradient explosion, and GIFStream contains flickering across the sequentially trained segments. 
% GIFStream trains sequentially using GOPs, causing error accumulation, inability to handle large motion, and flicker across segments. 
% These methods mainly target short sequences and simple motion.
In contrast, \fittername maintains higher and more consistent performance across all motion regimes, enabling high-fidelity free-viewpoint rendering with efficient memory usage.
As shown in Figure~\ref{fig:disocclusion},\ref{fig:teaser} and the supplementary material, \fittername produces sharper, more temporally coherent views. 
% Competing approaches struggle: 3DGStream shows drift under large motion, AT-GS produces floating artifacts and gradient explosion, and deformation-based models blur fine details—issues compounded by their high VRAM requirements that limit scalability to long sequences.

\vspace{-4mm}
\paragraph{Quantitative results.}
% \Cref{tab:qual_comp} further validate these observations. 
\Cref{tab:qual_comp} quantitatively demonstrates that \fittername outperforms all baselines in visual quality metrics across datasets. 
% \fittername consistently outperforms all the baselines in PSNR and other metrics for novel views across datasets. 
While 3DGStream, ATGS, and GIFStream support streaming longer sequences, they still yield suboptimal performance.
% Even streaming-based methods (3DGStream and AT-GS) yield suboptimal performance across all datasets and metrics. 
This demonstrates the effectiveness of our proposed representation and method for handling long-duration sequences, while maintaining compatibility with standard multimedia infrastructure.

% Leveraging UV-based pruning and the temporal fitting pipeline, \fittername effectively regularizes informative Gaussian primitives while maintaining compact memory footprints. 
% The scene compression strategy (Section~\ref{sec:scene_compression}) further enables practical modeling of arbitrary-length sequences.

\vspace{-3mm}
\paragraph{Long-term temporal consistency.}
To assess long-term temporal stability, we compute optical flow between consecutive timestamps from novel viewpoints. 
For methods that require a segmented training strategy, we compute optical flow for consecutive timestamps between segments. 
As shown in Figure~\ref{fig:optical_flow}, this demonstrates the inability of deformation based methods to maintain temporal consistency over time. 
% As shown in Figure~\ref{fig:optical_flow}, deformation-based methods are not consistent over time due to their inability to fit long sequences.
This further highlights the importance of online fitting methods like ours.
It is interesting to note that the rendering quality of 3DGStream and ATGS degrades over time whereas \fittername is consistent. 
% A 30 minutes long sequence is presented in the accompanying video which shows no loss in quality over time.

% bullet points for results
% - 3DGStream are struggle with large motion (dance figure in supp) and objects appearing/disappearing
% - 3dgstream performance degrade over time (gradient accumulation?)
% - atgs floaters in front of test cameras (for numbers)
% - because we have to split into segments for training, there is inconsistency between segments (flow figure)

\begin{table}[t]
\centering
\setlength{\tabcolsep}{0.95mm}
\renewcommand{\arraystretch}{1.2}
\caption{We present quantitative ablation study for various components of our method on PSNR, SSIM, and LPIPS.}
\vspace{-0.2cm}
\resizebox{0.98\columnwidth}{!}{
\begin{tabular}{l|ccc||l|ccc}
 \toprule
 \textbf{Method} & \textbf{PSNR}$^\uparrow$ & \textbf{SSIM}$^\uparrow$ & \textbf{LPIPS}$^\downarrow$ & \textbf{Method} & \textbf{PSNR}$^\uparrow$ & \textbf{SSIM}$^\uparrow$ & \textbf{LPIPS}$^\downarrow$ \\
 \midrule
 Ours & $27.41$ & $0.84$ & $0.28$ &         w/o Keyframe  & $20.95$ & $0.77$ & $0.38$ \\
 w/o UV Optim & $23.81$ & $0.79$ & $0.33$ & w/o Labeling & $25.42$ & $0.82$ & $0.31$ \\
 No Atlas & $27.43$ & $0.84$ & $0.28$ &     w/o Codec &  $27.41$  &  $0.84$ & $0.28$ \\
 No LPO & $27.52$ & $0.85$ & $0.27$ &       $-$ &  $-$  &  $-$ & $-$ \\

 \bottomrule
\end{tabular}}
\label{table:ablation} 
\vspace{-0.4cm}
\end{table}

\vspace{-1mm}
\subsection{Video Coding of \methodname}
\label{sec:video_coding}

We can encode \methodname using standard 2D video codecs while preserving quality of the 4D scene. 
Because \fittername maintains temporally consistent UV layouts and only updates per-pixel attributes over time for dynamic regions, the atlas sequence exhibits \emph{strong spatial locality} and \emph{temporal coherence}, allowing direct reuse of mature video coding pipelines.

For each frame in \methodname, we group PackUV layers into 8-bit triplets before encoding them with lossless codecs (\eg,~FFV1, HuffYUV) via FFmpeg~\cite{ffmpeg}. 
Each channel is globally normalized using per-channel min/max computed over the entire sequence. 
To guarantee bit-exact recovery of the original attributes, we transmit a compact sidecar containing the exact normalization parameters. 
Decoding is thus invertible: it recovers the atlas sequence, which is de-normalized to reproduce the original values.

This design reduces 4D Gaussian scenes to conventional video assets without custom codecs, enabling storage, streaming, and decoding with off‑the‑shelf tooling. 
In practice, we obtain a perfect reconstruction with zero error in the lossless setting via FFV1.
These results confirm that \methodname enables treating 4DGS as standard video content while retaining faithful scene recovery.

% Our objective is to encode the \methodname representation that encodes all 3DGS attributes into a compact 2D video that is both easy to store and can be converted back to a 4D scene representation via lossless video codecs (like h265, ffv1, etc.). 
% Given the \methodname structurally organizes scene attributes in an image-like format, and \fittername only optimizes the dynamic UV pixel over time while keeping the static temporally consistent, we can directly adopt video encoding pipelines.

% As a widely used video encoding and decoding tool, FFmpeg~\cite{ffmpeg} integrates various video codec protocols. 
% We get a compression ratio of xx with negligible degradation. 
% This further verifies that through our design, the 4DGS is able to be processed like a video.

\vspace{-1mm}
\subsection{Ablation Study and Discussion}
\label{sec:ablation}

We perform comprehensive ablations on \datasetname to validate each component of our method, examining atlas-based packing, Gaussian labeling, low-precision optimization (LPO), video coding, direct UV-space optimization, and video keyframing (Table~\ref{table:ablation}), using PSNR, SSIM, and LPIPS for evaluation.
Removing UV initialization and UV pruning (optimizing only via post-hoc UV projection) leads to a clear drop in PSNR and other metrics, confirming the importance of direct UV-space optimization for fine detail preservation.
We also assess mapping from uniform UV-space Gaussians to the \methodname atlas. 
Because deeper UV layers contain few primitives, compressing them into lower-resolution atlases incurs negligible quality loss.

Our ablations also demonstrate that low-precision optimization is effectively lossless and provides a superior alternative to post-training quantization for efficient Gaussian storage. Moreover, LPO enables easy integration with standard video-coding pipelines without requiring specialized post-processing steps such as in Motion Layering, VCubed, or GIFStream. 
Because our \methodname representation consists entirely of 8-bit images, it can be directly encoded as video and decoded back without reconstruction loss.
Finally, we assess the effect of video keyframing. Resetting gradients at keyframes helps maintain spatial and temporal consistency, preventing the gradual quality degradation observed when training without keyframes. 
Removing keyframing results in a clear decline in average PSNR and other metrics. Additional results are in the supplementary material.

% \section{Limitation and Conclusion}
% \label{sec:conclusion}

% While our method achieves high-quality FVV renderings, several limitations remain that warrant future exploration. 
% One key challenge stems from the inherent unstructured nature of 3D Gaussian representations. 
% Although our approach introduces structure through the UV projection, we do not currently enforce temporal consistency in this structure across frames. 
% Future work could explore regularizing these structured UV pixels to improve fitting stability and quality over long-duration dynamic sequences. 
% Additionally, making this representation compatible with AR/VR devices would be a compelling direction, enabling direct 4D streaming for immersive applications.

\vspace{-2mm}
\section{Conclusion}
\label{sec:conclusion}
\vspace{-1mm}
We presented \methodname, a unified 4D Gaussian representation that reorganizes volumetric video into structured UV atlases compatible with standard video codecs, and \fittername, an efficient fitting pipeline that maintains long-term temporal consistency under large motions and disocclusions. 
By directly optimizing all Gaussian attributes in the UV domain, our approach enables high-quality reconstruction for arbitrarily long sequences while improving streaming efficiency. 
We further introduced \datasetname, the largest long-duration multi-view 4D dataset to date, providing challenging benchmarks for future research. Extensive experiments demonstrate that our method outperforms prior work in quality, scalability, and practicality, offering a promising step toward deployable volumetric video in real-world systems.

\paragraph{ACKNOWLEDGEMENTS}
This research was supported by ONR DURIP grant N00014-23-1-2804, NSF CAREER award $\#$2143576, and an Amazon Cloud Credits award.
We also want to thank Emre Arslan for building the first version of the viewer. 
We also appreciate the time and efforts of all the participants for data collection.
% \vspace{-4mm}
{
    \small
    \bibliographystyle{ieeenat_fullname}
    \bibliography{main}
}

\clearpage
\setcounter{page}{1}
\maketitlesupplementary

\section{\datasetname Dataset}
We captured \datasetname, a novel real-world dataset featuring long-horizon, multi-view video sequences. 
\datasetname comprises \numsequences such sequences, with over \numimgs~(billion) frames in total, and encompasses a diverse array of scenarios, including human-human interaction, human-object interaction, robot-object interaction, among others. 
The sequences in \datasetname average approximately $10$ minutes in length, with some extending up to $30$ minutes.
Notably, to establish \datasetname as a comprehensive benchmark for evaluating dynamic reconstruction approaches, we have curated sequences of varying difficulty levels, aiming to cover a broad spectrum of real-world settings. 
For instance, in terms of motion speed, \datasetname includes movements ranging from slow actions like "inching along" to fast-paced activities such as playing basketball, thereby posing challenges for handling motion blur. 
Regarding motion scale, \datasetname captures both small-scale interactions, like table-top object manipulation, and large-scale activities, such as dance, pickle ball, volleyball, etc. 
Furthermore, \datasetname features diverse object categories, including rigid and articulated objects, as well as some reflective and transparent items.
To the best of our knowledge, \datasetname significantly surpasses existing datasets in its domain concerning sequence length, number of camera views, motion complexity, dynamic diversity, and overall data volume.
More details about the captured sequences are presented in Table~\ref{tab:supp_dataset}.
The Capture system is discussed in Supp. \cref{supp:sec:AVL}.

\begin{table}[h]
\centering
\setlength{\tabcolsep}{0.65mm}
\caption{PSNR comparison with the baselines on N3DV (\textit{flame\_salmon}), DeskGames, and Technicolor datasets.}
\vspace{-0.2cm}

\scalebox{0.78}{\begin{tabular}{l|c|c|c|c|c|c}
 \textbf{Dataset}       & Motion L. & 4DGS & Ex4DGS & 3DGStream & GIFStream & Ours \\
 \midrule
  \textbf{N3DV}         & \cellcolor{yellow!40}32.55  &  32.01  & \cellcolor{orange!40}32.11 & 31.67 & 28.42 & \cellcolor{red!40}33.06 \\
  \midrule
  \textbf{DeskGames}    & \cellcolor{yellow!40}30.89  &  30.11  & 29.48 & \cellcolor{orange!40}30.51 & 29.94 & \cellcolor{red!40}32.74 \\
  \midrule
  \textbf{TechniColor}  & \cellcolor{yellow!40}31.77  &  28.91  & \cellcolor{orange!40}31.33 & 25.48 & 27.42 & \cellcolor{red!40}31.87 \\
 \bottomrule
\end{tabular}}
\label{table:rebut_compare} 
\vspace{-0.2cm}
\end{table}

\section{Additional Qualitative and Quantitative Results}

\begin{figure*}[!h]
\vspace{-6mm}
\centering
\includegraphics[width=5.4in]{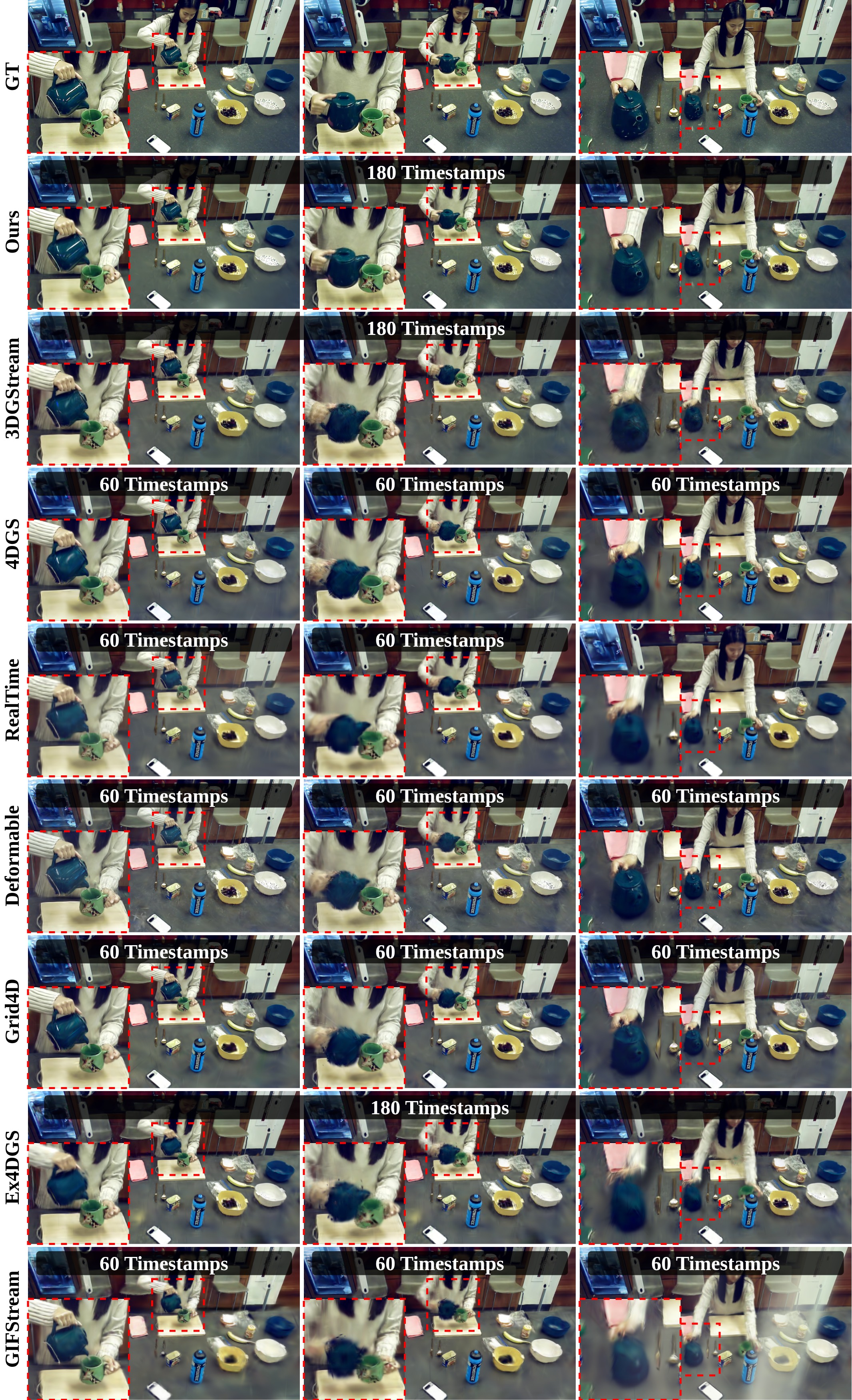} 
\caption{
Baseline comparison on \datasetname's \textit{Kitchen} sequence.
}
\label{fig:supp_kitchen_results}
\end{figure*}

More qualitative results are shown in figures 
\ref{fig:supp_kitchen_results},
\ref{fig:supp_baby_result},  \ref{fig:supp_baselines_selfcap},  \ref{fig:supp_spot_result} at the end of supplementary. From the results, it can be seen that our method, \methodname, significantly outperforms other methods in reconstructing large motions and disocclusions while maintaining the overall quality.

We also evaluate on three additional datasets, including one more sequence from N3DV, Technicolor, and the DeskGames dataset from Motion Layering.
A summarized comparison is shown in \cref{table:rebut_compare}.
Our method performs better across datasets.
Metrics are collected from Motion Layering except ours and GIFStream for the same experimental setup.

\section{Lossy Post-Optimization UV Mapping for Scenes}

\begin{figure*}[!h]
\centering
\includegraphics[width=6.8in]{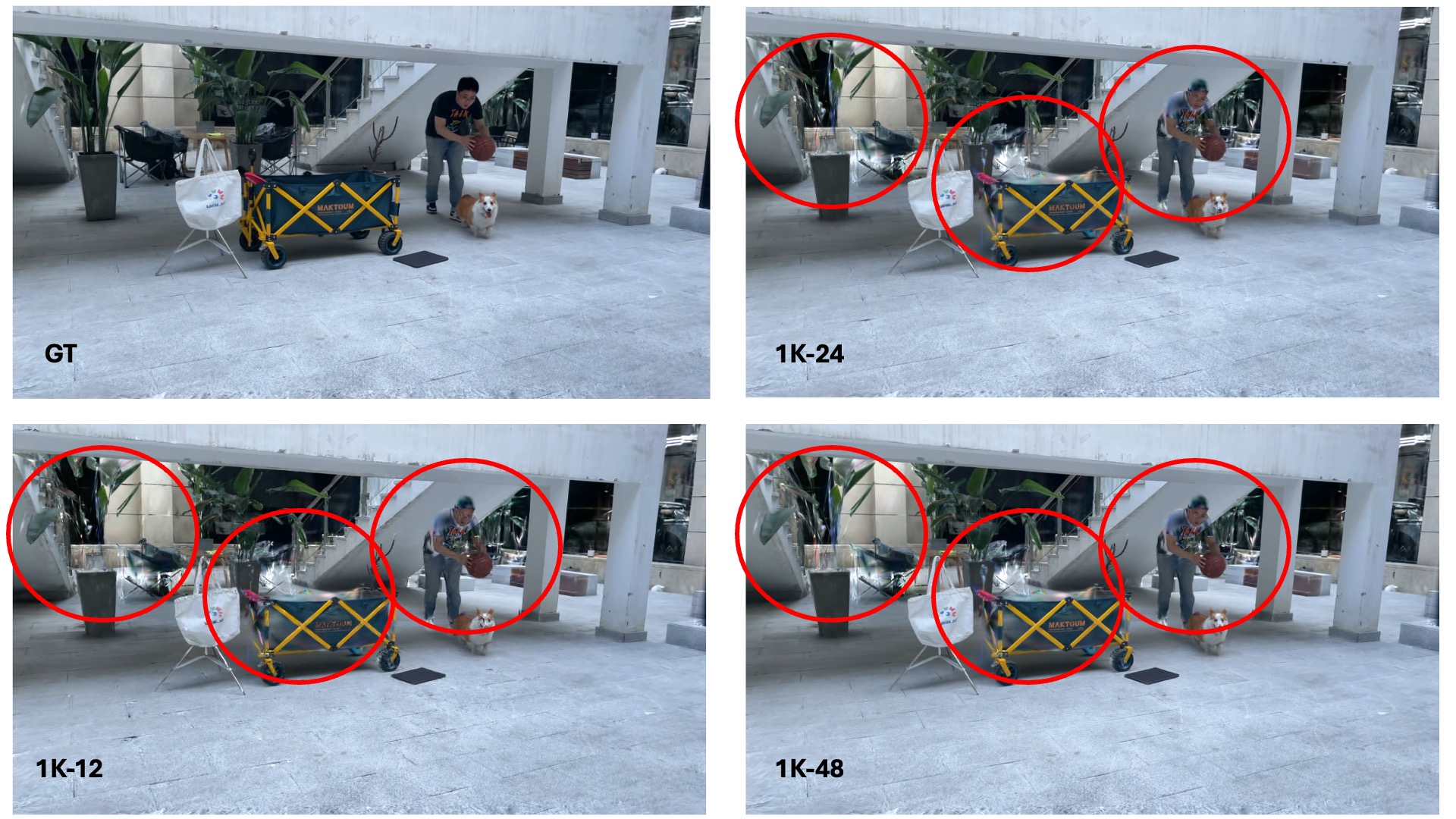} 
\caption{
Post-optimization lossy UV mapping fails to capture details of a real-world scene even with 48 layers and 1K resolution.
}
\label{fig:supp_uvgs_mapping}
\end{figure*}

Fig.~\ref{fig:supp_uvgs_mapping} show lossy reconstruction via UVGS~\cite{rai2025uvgs} mapping for a real-world scene. Even when using 48 layers and 1K resolution of UV maps, there are missing Gaussians resulting in clearly visible artifacts in the renderings.
This clearly illustrates the importance of UV mapping during optimizing the Gaussians as done in \methodname.

\begin{table}[h]
\centering
\setlength{\tabcolsep}{0.65mm}
\caption{Storage comparison with the baselines (30 frames).}
\vspace{-0.1cm}

\scalebox{0.82}{\begin{tabular}{l|c|c|c|c|c|c}
 \textbf{Method}         & LG   & Grid4D & Ex4DGS & 3DGStream & GIFStream & Ours \\
 \midrule
  \textbf{Storage (MB)}  & 950  &  \cellcolor{yellow!40}76    & 108    & 204       & \cellcolor{orange!40}16        & \cellcolor{red!40}10 \\
 \bottomrule
\end{tabular}}
\label{table:rebut_baseline_storage} 
\vspace{-0.3cm}
\end{table}

\begin{figure*}[!h]
\centering
\includegraphics[width=6.8in]{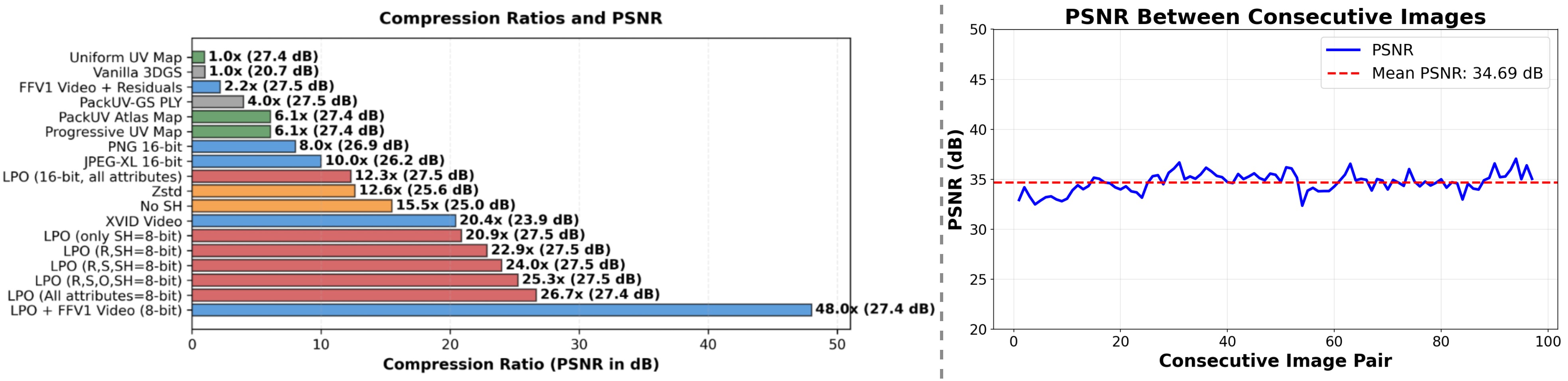} 
\caption{
(Left) Compression evaluation via different methods. 
(Right) PSNR consistency over time.
}
\label{fig:rebutt_compression}
\end{figure*}

\section{Video Coding and \methodname Storage}
\label{sec:scene_compression}

\methodname allow easy and \textbf{lossless} encoding of Volumetric videos using standard 2D video codecs while preserving quality of the 4D scene. 
Because \fittername maintains temporally consistent UV layouts and only updates per-pixel attributes over time for dynamic regions, the atlas sequence exhibits \emph{strong spatial locality} and \emph{temporal coherence}, allowing direct reuse of mature video coding pipelines.
This solved one of the major drawbacks of streaming based methods like GIFStream, 3DGStream and AT-GS - high storage requirements of the trained models for every timestamp for lossless conversion. 
Using the FFV1 codec for lossless video conversion is giving us an average storage rate of under \textbf{10 MBPS}.
\cref{table:rebut_baseline_storage} compares storage with SOTA 4DGS methods. 
Our method outperforms all volumetric video streaming baselines and significantly surpasses specialized per-frame compression methods such as LG~\cite{fan2024lightgaussian}.

We analyze the compression of PackUV atlases via different techniques including quantization of individual 3DGS attributes, using different video and image compression techniques, and mix of both. 
The results are presented in \cref{fig:rebutt_compression}(left). 
FFV1 lossless compression with 8-bit LPO achieves the highest compression ratio with only a \textbf{0.11\,dB} PSNR drop.

Another advantage of \methodname is it's structural arrangement, which can be used for mapping the scene to a latent space to achieve neural compression~\cite{rai2025uvgs, xiang2025repurposing}.
Unlike UVGS~\cite{rai2025uvgs}, GaussianAtlas~\cite{xiang2025repurposing}, \methodname allows us to represent the entire scene into a single multiscale atlas.
This further reduces the complexity of neural networks required to compress it to a latent space.
The objective is to compress a single-layer UV atlas representation—which encodes full \methodname atlas to a compact representation that is both easy to store and invertible back to a 4D scene representation. Since the UV format structurally organizes scene attributes in an image-like format, we directly adopt image-based compression pipelines.

\subsection{Neural Compression}
Neural Compression is a widely used technique for compressing various modalities~\cite{egosonics, rai2025uvgs}.
We consider a neural network architecture for compressing \methodname inspired from UVGS~\cite{rai2025uvgs}. 
We follow the designing by using three lightweight encoder-decoder networks, each responsible for compressing a specific modality:
\begin{itemize}
    \item \textbf{Position Encoder-Decoder:} maps spatial information \( \sigma_i \in \mathbb{R}^{3 \times H_0 \times W_0} \)
    \item \textbf{Appearance Encoder-Decoder:} maps RGB/SH color and opacity \( \{c_i, o_i\} \in \mathbb{R}^{4-45 \times H_0 \times W_0} \)
    \item \textbf{Covariance Encoder-Decoder:} maps scale and rotation \( \{s_i, r_i\} \in \mathbb{R}^{7 \times H_0 \times W_0} \)
\end{itemize}

Each modality is encoded into a 3-channel latent image per frame. Let \( L^t \in \mathbb{R}^{M \times N \times 3} \) be the per-frame latent representation at time \( t \). These latent images are stored along with the shared decoder weights for the entire scene. This approach offers a consistent compression ratio ranging from \textbf{[8:1]} to  \textbf{[128:1]} at the cost of minimal degradation (\textbf{PSNR drop: [1.5] dB} to \textbf{PSNR drop: [4.5] dB}). And the decoding process is very efficient due to the light-weight MLP based decoder.

\begin{table*}[!h] % h: here, t: top, b: bottom, p: page of floats
  \centering 
  \fontsize{8pt}{10pt}\selectfont
  % \vspace{-0.3cm}
  \caption{\textbf{\datasetname Dataset Layout.}
We present our newly captured dataset \datasetname sequences . We report number of sequences captured, total timestamps (T), total cameras used to capture the sequence, FPS, setting type, and view range. 
We also report special tags for each dataset describing the type of activity in the captured sequence.
\datasetname contains \numsequences diverse sequences totaling over \numimgs~(billion) high-quality frames, recorded with more than 50 cameras at $1920 \times 1200$ resolution.
The capture system supports up to 90 FPS, providing high temporal fidelity.
Tags: 
RI - Robot Interaction, 
HI - Human-human Interaction, 
OI - Object Interaction, 
SP - Sports, 
LM - Large Motion, 
DO - Disocclusion, 
TR - Transparent/Reflective Objects,
EN - Entertainment
.
% \rai{in progress...}
}
  \resizebox{\linewidth}{!}{
  \begin{tabular}{@{}L{2cm}C{1.2cm}C{1.0cm}C{1.5cm}C{1.5cm}C{2cm}C{1.0cm}@{}}
\toprule
Sequence & Num Sequences & Num Cameras & FPS & Setting & Tags & View Range \\
\midrule

Baby Dance
& 3
& 88
& 30
& Studio
& EN, HI, DO
& 360 \\

Spot
& 3
& 88
& 30, 90
& Studio
& RI, HI, OI, DO
& 360 \\

Volleyball
& 3
& 88
& 30, 90
& Studio
& SP, HI, LM, DO
& 360 \\

Kitchen
& 3
& 55
& 30
& Non-studio
& OI
& 320 \\

Woodwork
& 4
& 55
& 30
& Non-studio
& OI
& 320 \\

Pickleball
& 30
& 55
& 30
& Non-studio
& SP, HI, LM
& 300 \\

Meat Shop
& 11
& 84
& 30
& Non-studio
& OI, DO
& 320 \\

Kuka Robot
& 2
& 84
& 30
& Non-studio
& RI, OI
& 300 \\

Panda Robot
& 3
& 84
& 30
& Non-studio
& RI, OI
& 300 \\

Articulation
& 2
& 84
& 30
& Studio
& HI, OI, DO
& 360 \\

Chair Play
& 1
& 86
& 30
& Studio
& HI, OI, DO, LM
& 360 \\

Object Placement
& 1
& 88
& 30
& Studio
& HI, OI, DO, LM
& 360 \\

Dance 01
& 3
& 80
& 30
& Studio
& HI, DO, LM, EN
& 360 \\

Dance 02
& 7
& 84
& 30, 60, 90
& Studio
& HI, DO, LM, EN
& 360 \\

Dance 03
& 3
& 82
& 30, 60, 90
& Studio
& DO, LM, EN
& 360 \\

Dance 04
& 3
& 85
& 30
& Studio
& DO, LM, EN
& 360 \\

Yoga
& 3
& 78
& 30
& Studio
& LM, SP
& 360 \\

Tools Play
& 2
& 82
& 30
& Studio
& LM, DO, OI
& 360 \\

Board Games
& 4
& 86
& 30
& Studio
& HI, OI, LM, DO
& 360 \\

Photography
& 4
& 88
& 30
& Studio
& HI, OI, LM, DO
& 360 \\

Conversation
& 5
& 84
& 30
& Studio
& HI, LM, DO
& 360 \\
\midrule

\textbf{\datasetname (TOTAL)}
& \textbf{100}
& \textbf{55--88}
& 30-90
& -
& -
& $\sim$\textbf{360} \\

\bottomrule
\end{tabular}
    }
  \label{tab:supp_dataset} 

\end{table*}

\section{\fittername Optimization Details}

\subsection{Gaussian Labeling:} 

We use RAFT~\cite{teed2020raft} is used to estimate the optical flow for each set of images.
Flow masks are used to label the Gaussians as dynamic or static and only dynamic Gaussians are optimized during the training. 
We implemented a custom CUDA kernel to expedite this. 
The process is Explained in Algorithm~\ref{alg:cuda_flow_masking}.
As a fallback mechanism, we additionally support a simple point-projection strategy that relies solely on the Gaussian centers, without accounting for their full covariance. 
In the setting, even when the flow-based masking module is disabled, we observe that training the sequence using a keyframing strategy remains stable and temporally coherent. 
Specifically, the 3D Gaussians for each new keyframe are initialized using the optimized Gaussian parameters from the preceding keyframe. 
The same is done for the transition frames.
This warm-start initialization propagates geometry and appearance across time, effectively enforcing temporal continuity and preserving structural consistency between consecutive segments, despite the absence of explicit motion-aware supervision.
We believe, this is possible by fact that PackUV focus learning on surface-relevant regions and most of them remain static during training of the subsequent timestamps.

\begin{algorithm*}[t]
\caption{Covariance-Aware Flow Masking with CUDA Acceleration}
\label{alg:cuda_flow_masking}
\begin{algorithmic}[1]
\Require Gaussian parameters $\{\boldsymbol{\mu}_i, \mathbf{s}_i, \mathbf{q}_i\}_{i=1}^N$, camera views $\mathcal{C}$, flow masks $\{M^c\}_{c \in \mathcal{C}}$

\Ensure Dynamic mask $D_i \in \{0,1\}$ for each Gaussian $i$

\State Initialize $D_i \leftarrow 0$ for all $i = 1, \ldots, N$

\For{each camera $c \in \mathcal{C}$}
    \State \textbf{CUDA Kernel (parallel over all Gaussians):}
    \For{each Gaussian $i$ \textbf{in parallel}}
    
        \State \textbf{// 1. Compute 3D covariance and transform to camera space}
        \State $\Sigma^{3D}_i \leftarrow \mathbf{R}(\mathbf{q}_i) \cdot \text{diag}(\mathbf{s}_i^2) \cdot \mathbf{R}(\mathbf{q}_i)^\top$
        \State $\Sigma^{3D}_{i,\text{cam}} \leftarrow \mathbf{T}_c \cdot \Sigma^{3D}_i \cdot \mathbf{T}_c^\top$
        \State $\boldsymbol{\mu}_{i,\text{cam}} \leftarrow \mathbf{T}_c \cdot \boldsymbol{\mu}_i$

        \hfill
        \State \textbf{// 2. Project to 2D using EWA splatting}
        \State Compute Jacobian: $\mathbf{J} = \begin{bmatrix} f_x/z & 0 & -f_x x/z^2 \\ 0 & f_y/z & -f_y y/z^2 \end{bmatrix}$ at $\boldsymbol{\mu}_{i,\text{cam}}$
        \State $\Sigma^{2D}_{i,c} \leftarrow \mathbf{J} \cdot \Sigma^{3D}_{i,\text{cam}} \cdot \mathbf{J}^\top$
        \State $\Sigma^{2D}_{i,c}[0,0] \leftarrow \Sigma^{2D}_{i,c}[0,0] + 0.3$ \Comment{Low-pass filter}
        \State $\Sigma^{2D}_{i,c}[1,1] \leftarrow \Sigma^{2D}_{i,c}[1,1] + 0.3$
        \State $\mathbf{m}_{i,c} \leftarrow \text{NDC2Pixel}(\mathbf{P}_c \cdot \boldsymbol{\mu}_i)$ \Comment{Project mean to pixels}

        \hfill
        \State \textbf{// 3. Check covariance validity}
        \State $\text{det} \leftarrow \Sigma^{2D}_{i,c}[0,0] \cdot \Sigma^{2D}_{i,c}[1,1] - \Sigma^{2D}_{i,c}[0,1]^2$
        \If{$\text{det} \le 10^{-7}$ \textbf{or} $\text{det} < 0$ \textbf{or} $z \le 0$}
            \State $D_{i,c} \leftarrow 0$ \Comment{Invalid, mark static}
            \State \textbf{continue}
        \EndIf

        \hfill
        \State \textbf{// 4. Compute sampling radius from eigenvalues}
        \State $\text{tr} \leftarrow \Sigma^{2D}_{i,c}[0,0] + \Sigma^{2D}_{i,c}[1,1]$
        \State $\lambda_{\max} \leftarrow \frac{\text{tr} + \sqrt{\max(0, \text{tr}^2 - 4 \cdot \text{det})}}{2}$
        \State $r \leftarrow \min\left(r_{\max}, \left\lceil 3\sqrt{\lambda_{\max}} \right\rceil\right)$ \Comment{3-sigma, clamped}

        \hfill
        \State \textbf{// 5. Test ellipse overlap with flow mask}
        \State $D_{i,c} \leftarrow 0$
        \For{$\mathbf{p} \in \{\mathbf{m}_{i,c} + \boldsymbol{\delta} \mid \|\boldsymbol{\delta}\|_\infty \le r\}$}
            \If{$\mathbf{p}$ within image bounds}
                \State $d^2 \leftarrow (\mathbf{p} - \mathbf{m}_{i,c})^\top (\Sigma^{2D}_{i,c})^{-1} (\mathbf{p} - \mathbf{m}_{i,c})$
                \If{$d^2 \le 9$ \textbf{and} $M^c(\mathbf{p}) > 0.5$}
                    \State $D_{i,c} \leftarrow 1$
                    \State \textbf{break}
                \EndIf
            \EndIf
        \EndFor
    \EndFor

    \hfill
    \State \textbf{// Aggregate on host (OR across cameras)}
    \For{each Gaussian $i$}
        \State $D_i \leftarrow D_i \lor D_{i,c}$
    \EndFor
\EndFor

\State \Return $\{D_i\}_{i=1}^N$
\end{algorithmic}
\end{algorithm*}

\subsection{Gradient thresholding:} 
% One common problem with most of the existing dynamic reconstruction methods is the progressive growth in the number of Gaussians over time. 
% This results in OOM and other issues during fitting longer sequences.
% To avoid such growth of Gaussians over time, we introduce a gradient thresholding mechanism that caps the maximum number of points that can be used to represent a scene. 
% It removes the Gaussians that do not significantly contribute to the scene representation, while keeping the ones that do using \todo{using what? describe here}.
% REFINED VERSION BELOW:
A common challenge in the existing dynamic 3D reconstruction methods is the uncontrolled growth of the number of Gaussian primitives over time.
This progressive accumulation leads to excessive memory consumption and frequent out-of-memory (OOM) errors, especially when fitting long video sequences.
To mitigate this issue, we introduce a \textit{gradient thresholding} mechanism that constrains the total number of Gaussians allowed during optimization. 
The key idea is to retain only those Gaussians that meaningfully contribute to the reconstruction objective while pruning away redundant or inactive ones.

Let the loss function be denoted by \( \mathcal{L} \), and consider a Gaussian primitive \( g_i \). We compute the gradient norm of its contribution to the loss:
\[
\|\nabla_{g_i} \mathcal{L}\|_2
\]
If this norm falls below a predefined threshold \( \tau \), the Gaussian is considered non-contributory and is removed from the scene representation. Formally:
\[
\text{If } \|\nabla_{g_i} \mathcal{L}\|_2 > \tau, \quad \text{then } g_i \text{ is pruned}
\]

This strategy ensures that only actively contributing Gaussians are retained during optimization, effectively capping the total number of primitives and maintaining computational efficiency. 
By doing so, we avoid unnecessary memory overhead and preserve high-quality reconstruction over long temporal horizons.

\subsection{Video Keyframing}

To efficiently fit given synchronized multi-view videos with $T$ frames, represented as a set $\{V_n\}_{n=1}^N$, where N is the total number of views.
We divide each video into s set of $m$ temporal segments.
To do so, for each frame $t$, we compute the optical-flow magnitude $M(t)$ on one video, select the top $(m-1)$ magnitude peaks with a minimum separation $\theta$, and use the first frame of every segment as a keyframe. 
% \srinath{Shoudn't keyframing be done before diving into temporal segments?}
% \[F^K_0 = 1, \qquad\] 
% \[ \{F^K_i\}_{i=1}^{m-1} = \operatorname{Top}_{m-1}\big(M(t)\big) \ \text{s.t.}\ |F^K_i - F^K_{i-1}| \ge \tau . \]
% \srinath{Notation above is too cumbersome, please simplify.}
These keyframes define the segment boundaries. 
For each keyframe $F^K_i$, the PackUV Gaussians are initialized from the previous keyframe which preserves temporal and spatial consistency.
The frames between keyframes are treated as transition frames. 
Each transition frame $F^t$ is initialized from the preceding frame and refined with a few training iterations compared to $F^K$:
\vspace{-2mm}
\[
\mathcal{G}(K) \leftarrow \text{Update}\!\left(\mathcal{G}(K-1)\right),
\mathcal{G}(t) \leftarrow \text{Update}\!\left(\mathcal{G}(t-1)\right).
\]

% \vspace{-2mm}
This staged, stream-based strategy enables efficient reconstruction of high-fidelity dynamic scenes while allowing us to parallelize the fitting process.
Frames exhibiting high drift, occlusions/disocclusions, or appearance breaks are promoted to keyframes.
This keyframing technique helps us handle arbitrary length sequences, large motions, and disocclusions without quality degradation over time.

Through our experiments we observe that keeping keyframes farther apart can cause the quality to degrade over time, thus we keep the keyframe threshold to 30 timestamps.

\subsection{Low-Precision Training}

Unlike most prior works that employ lossy quantization after optimization for storage efficiency, \fittername utilizes low precision optimization (LPO) for learning the Gaussian attributes 
$\theta = \{\mathbf{\mu}, \mathbf{s}, \mathbf{r}, \alpha, \mathbf{c}\}$.
Our experiments show that LPO effectively compensates for the quantization loss, maintaining both PSNR and training efficiency.
At each iteration, the renderer operates on a quantized proxy $\tilde{\theta}$ obtained by a uniform $K$-bit fake quantizer with scale $\Delta$.
We use a straight-through estimator (STE) to preserve gradient flow while keeping master weights in FP32. 
The training objective remains the same, combining an L1+SSIM photometric term with a scheduled regularizers.
Backpropagation updates FP32 master parameters with Sparse Adam optimizer~\cite{taming3dgs, albedogan}.

We use 8-bit reduced precision for $\{\mathbf{s}, \mathbf{r}, \alpha, \mathbf{c}\}$ and 16-bit for $\{\mathbf{x}\}$. 
During storage, the 16-bit $\{\mathbf{x}\}$ values are split into two 8-bit parts. 
This 8-bit-per-channel design makes \methodname readily compatible existing video coding infrastructures, enabling the direct application of both lossless or lossy compression methods (\eg,~HEVC, AVC, FFV1).
We also conduct several experiments to evaluate the loss incurred from reduced-precision training using different levels of bit quantization. 
From these experiments, we observe that that quantizing the position parameters ($\{\mathbf{x}\}$) cause the largest PSNR drop during training. 
Consequently, we retain them in FP16, which introduces negligible error during optimization. 
In contrast, the other attributes ($\{\mathbf{s}, \mathbf{r}, \alpha, \mathbf{c}\}$) show no noticeable PSNR drop in our experiments, even when quantized to 8-bit integers after appropriate scaling.

\section{Viewer}
% We also built a custom viewer for rendering our \methodname atlas maps on top of the Tiny Gaussian Splatting Viewer by Li Ma, which originally supports OpenGL and CUDA backends for static scenes. Our viewer extends this functionality to enable interactive camera control, real-time playback at arbitrary frame rates, and frame-by-frame navigation through dynamic sequences.

% Rendering is performed using OpenGL shaders and shader storage buffer objects (SSBOs). Each video frame is represented as a \methodname atals stored as a npz file, decreasing our loading latency compared to .ply files. And to accelerate playback, we cache a flattened, OpenGL-ready version of each frame on the CPU. At runtime, when advancing to a new frame, the viewer copies this preprocessed buffer to the GPU for rendering. Between frame updates, the current buffer remains active, allowing continuous rendering and smooth scene exploration.

% New viewer begins from here:

The viewer renders a dynamic 4D Gaussian scene as a time-ordered sequence of Gaussian frames. At time step \(t_k\), the scene is represented as
\[
\mathcal{G}(t_k)=\{g_i^{(k)}\}_{i=1}^{N_k},
\]
where each Gaussian \(g_i^{(k)}\) stores a 3D mean \(\bm{\mu}_i\), covariance \(\bm{\Sigma}_i\), opacity \(\alpha_i\), and appearance parameters (e.g., SH coefficients). A stream manager loads PackUV-encoded frame data sequentially, maintains a small look-ahead cache, and exposes both CPU-side Gaussian data and GPU-ready packed metadata for the current frame.

During playback, the active frame index is advanced according to the target frame rate,
\[
k(t)=\lfloor ft \rfloor \bmod N,
\]
where \(f\) is the playback FPS and \(N\) is the number of frames. When a new frame is selected, its packed representation is uploaded to the renderer and the Gaussian set is updated before drawing.

Rendering follows the standard 3D Gaussian splatting pipeline. A Gaussian in world coordinates is projected into screen space using the current camera pose and intrinsics. For a 3D point \(\bm{X}\), projection is given by
\[
\bm{X}_c=\bm{R}\bm{X}+\bm{t}, \qquad
\tilde{\bm{x}}=\bm{K}\bm{X}_c,
\]
with image coordinates obtained by perspective division. The projected 2D covariance is approximated as
\[
\bm{\Sigma}_{2D}=\bm{J}\bm{W}\bm{\Sigma}_{3D}\bm{W}^\top \bm{J}^\top,
\]
where \(\bm{W}\) is the local world-to-camera transform and \(\bm{J}\) is the Jacobian of the projection function. The resulting ellipse defines the screen-space footprint of each splat.

To support correct transparency, Gaussians are depth-sorted with respect to the current camera and composited back-to-front:
\[
\bm{C}=\sum_{i=1}^{M} T_i \alpha_i \bm{c}_i,
\qquad
T_i=\prod_{j<i}(1-\alpha_j),
\]
where \(\bm{c}_i\) is the color contribution of the \(i\)-th splat. 
The viewer additionally supports multiple shading modes, including depth visualization and spherical harmonics shading, making it suitable for interactive inspection of dynamic PackUV-based 4DGS sequences.

Excluding I/O, our viewer achieves $>\!200$ FPS; including I/O, it maintains $\ge\!30$ FPS on a NVIDIA 4090 GPU with a set number of GS in the range of 300K-500K for every scene. 
The viewer's rendering performance independent of sequence length due to buffering.

% \begin{figure*}[!ht]
% \centering
% \includegraphics[width=6.6in]{images/method_v4.jpg} 
% \caption{
% Method.
% }
% \label{fig:supp_method_supp}
% \end{figure*}

\section{CAPTURE System and Camera Synchronization}
\label{supp:sec:AVL}

\begin{figure}[ht]
    \centering
    \includegraphics[width=0.9\linewidth]{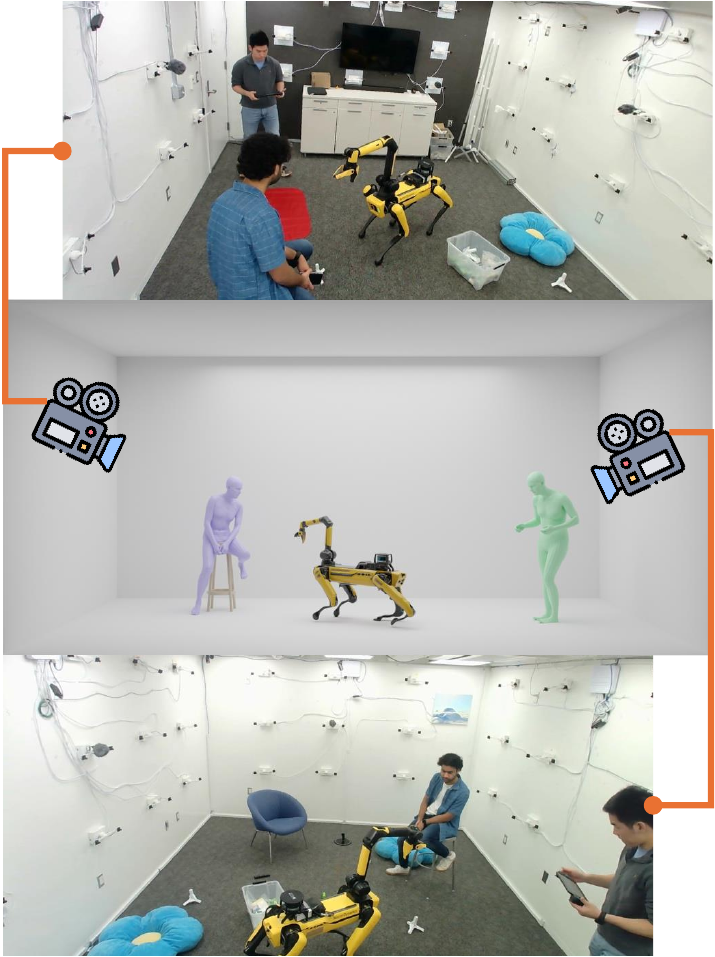}
    \caption{\textbf{CAPTURE Studio Layout.}}
    \label{fig:supp_brics}
\end{figure}

For capturing the sequences in \datasetname, we constructed a dedicated capture studio equipped with $88$ synchronized static cameras, as shown in Figure~\ref{fig:supp_brics}.
For the non-studio captures, we also built a wireless version of the similar capture setup.
The ultra-large scale of the \datasetname dataset raises great challenges to the corresponding data processing steps.
To this end, we develop an automatic pipeline to handle camera calibration, color and lighting correction, and automatic synchronization.
Notably, the automatic synchronization is efficiently achieved by a carefully designed data structure in Section~\ref{supp:sec:AVL}. 
These cameras are uniformly distributed across the four walls of a rectangular room, enabling the capture of fine-grained details with minimal occlusion. 
\datasetname provides high-resolution frames ($1920\times 1200$) at frame rates of $30$, $60$, or $90$ FPS, selected based on the level of dynamics in each sequence.
We plan to make \datasetname publicly available, with the goal of establishing it as a new benchmark for evaluating general-purpose, long-horizon dynamic reconstruction.

\begin{figure*}[!ht]
\centering
\includegraphics[width=6.6in]{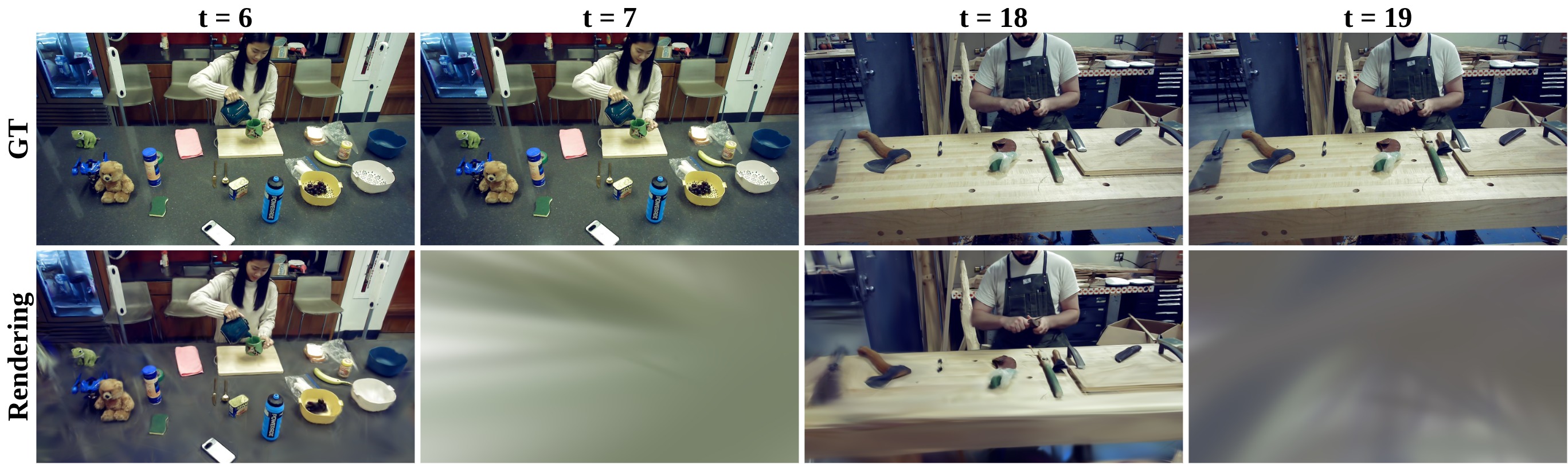} 
\caption{
Shows gradient explosion in ATGS training.
}
\label{fig:grad_explode}
\end{figure*}

\paragraph{CAPTURE} is a designated multi-view capture system with $88$ cameras, suitable for various kinds of capture settings.
This section provides an overview of an AVL tree implementation designed specifically for managing data captured by this system.
Based on the timecode of each frame data, this implementation achieves automatic synchronization and efficient search. 
In practice, it is non-trivial and time-consuming to achieve high-accuracy synchronization efficiently when working with such a large number of cameras.
We introduces an AVL tree implementation designed specifically for automatically synchronizing, managing and querying frame data based on timecodes. 
AVL trees are self-balancing binary search trees, ensuring that operations like insertion, deletion, and search can be performed efficiently, typically in $O(logn)$ time, where $n$ is the number of nodes in the tree.

The primary goal of this AVL tree is to transform unstructured raw frame data into an efficient data structure with synchronized frames.
This structure allows for:
\begin{enumerate}
    \item Quick Lookups: Rapidly searching all the frames at some timecode within a specified tolerance (threshold).
    \item Data Persistence: Once an AVL tree is built, it can be saved as a binary file to avoid the need to rebuild it from the raw file in the future.

\end{enumerate}

\subsection{Implementation}
The overview of the implementation can be divided into two steps:
\begin{enumerate}
    \item Build an AVL tree for each camera;
    \item After randomly selecting a reference camera, iterate through all the frames in the reference camera, and search the closest frames from all the other AVL trees as the synchronized frames.
\end{enumerate}

\subsection{Build an AVL tree}
Following~\ref{alg:buildAVLTree}, We build an AVL tree for each camera based on the camera information file which stores the correspondence of the frame index $idx_i$ and the timecode $t_i$.
It takes as input each pair of $\{ idx_i, t_i\}$, and then inserts it as a node into the AVL tree. 
The AVL tree's ensures an insertion logic that the tree remains balanced after each new node is added.

\begin{algorithm}
\caption{Build an AVL Tree from a Camera Information File}
\label{alg:buildAVLTree}
\begin{algorithmic}[1] % The [1] enables line numbers
\Function{BuildAVLTree}{filename}
    \State $\text{root node } r \gets \text{NULL}$
    \State Open a camera information file as $\mathcal{F}$
    \ForAll{$f$ in $\mathcal{F}$}
        \State Parse $f$ to get a $\{\text{timecode string }t_i , \text{frame index } idx_i$
        \State $r \gets \Call{InsertIntoAVL}{r, t_i, idx_i}$
        \Comment{InsertIntoAVL handles node creation and tree balancing}
    \EndFor
    \State Close $f$
    \State \Return $r$
\EndFunction
\end{algorithmic}
\end{algorithm}

With a built AVL tree, we can achieve efficient and fast lookups.
Since all the data is stored in an AVL tree, which is a type of Binary Search Tree (BST), nodes in a BST are organized such that all nodes in the left subtree of a node have timecodes less than the node's timecode, and all nodes in the right subtree have timecodes greater. 
This structure allows for efficient searching.
Given a timecode to search, the code iterates through the tree, keeping track of the node it has encountered so far whose timecode is closest to the target timecode. 
It also considers a threshold to ensure that the "closest" frame found is ``close enough".
If the difference of timecodes between the searched frame and reference frame is larger than a pre-defined threshold, that searched frame will be dropped and viewed as ``no synchronized frame".

\begin{algorithm*}
\caption{Find Closest Frame by Timecode}
\label{alg:findClosest}
\begin{algorithmic}[1] % The [1] enables line numbers
\Function{FindClosest}{root\_node $r$, target\_timecode $t$, threshold $\tau$}
    \State closest\_node $\hat{r} \gets \text{NULL}$
    \State min\_difference $d_{min} \gets \infty$
    \State current\_node $r' \gets $ root\_node $r'$

    \While{$r' \text{ IS NOT NULL}$}
        \State difference $d \gets |\text{current\_node.timecode} - \text{target\_timecode}|$
        \If{ $d < d_{min}$}
            \State $d_{min} \gets d$
            \State $\hat{r} \gets r'$
        \EndIf

        \If{$d = 0$}
            \State \textbf{break} \Comment{Exact match found, cannot be closer}
        \EndIf

        \If{$t < $ the timecode of $r'$}
            \State $r' \gets $ the left child node of $r'$
        \Else
            \State $r' \gets $ the right child node of $r'$
        \EndIf
    \EndWhile

    \If{$\hat{r} \text{ IS NOT NULL}$ \textbf{and} $d_{min} > \tau$}
        \State \Return $\text{NULL}$ \Comment{No node found within threshold}
    \Else
        \State \Return $\hat{r}$
    \EndIf
\EndFunction
\end{algorithmic}
\end{algorithm*}

\begin{algorithm*}
\caption{Synchronization Workflow}
\label{alg:Synchronization_Workflow}
\begin{algorithmic}[1]
    \Procedure{\phase{SynchronizeAndExtractFrames}}{}
        \State Define \varr{SyncMetaFile} path.

        \If{file at \varr{SyncMetaFile} exists}
            \State \varr{SyncInfo} $\gets$ \Call{\phase{LoadSyncInfo}}{\varr{SyncMetaFile}}
            \Comment{Loaded existing synchronization info.}
        \Else
            \State \varr{AVLTrees} $\gets$ \Call{\phase{GenerateAVLTrees}}{}
            \Comment{Leverage Algorithm~\ref{alg:buildAVLTree}}

            \State \varr{SyncInfo} $\gets$ \Call{\phase{SearchFramesUsingAVLTrees}}{}
            \Comment{Leverage Algorithm~\ref{alg:findClosest}}
            \State \Call{\phase{SaveSyncInfo}}{\varr{SyncInfo}, \varr{SyncMetaFile}}
        \EndIf

        \State \Call{\phase{ExtractAndSaveImageFramesFromVideos}}{\varr{SyncInfo}}
        \Comment{Extract specific frames directly from videos based on \href{https://github.com/pytorch/torchcodec}{torchcodec}.}
    \EndProcedure
\end{algorithmic}
\end{algorithm*}

\subsection{Iteratively Synchronize All the Cameras}
After building AVL trees for all the cameras, we prepare an automatic workflow (Algorithm~\ref{alg:Synchronization_Workflow}) to synchronize all of them.
Firstly, we sample a reference camera, either randomly or intentionally.
Then, we go through all the frames in the reference camera and look up the closest frame of every AVL tree as the synchronized frame.
Finally, we save the synchronization information as a json file for future use.
Moreover, based on \href{https://github.com/pytorch/torchcodec}{torchcodec}, we achieve efficiently extracting any specific frames directly from raw MP4 video file without the need to extracting all the frames first.

\section{Limitations and Future Work}
\label{sec:limitation}

While our method achieves high-quality 4D neural video renderings, some limitations remain that requires further exploration. 
A primary challenge arises from the inherently unstructured nature of 3D Gaussian representations. 
Although our approach introduces structure through UV projection, it still requires enforcing a large spatial arrangement to capture the fine details of real-world scenes. 

\methodname can sometimes produce dragged Gaussian artifacts when the optical flow is estimated inaccurately, which becomes a bottleneck in the pipeline. Although setting more keyframes and reducing optical flow threshold could fix the problem, improving the flow prediction module or adopting a more robust mapping strategy could lead to better temporal consistency.

Furthermore, while \methodname enables the use of video coding infrastructure by mapping to a single frame, discovering more optimal mappings could further improve storage efficiency. 
Another promising direction is to make this representation compatible with AR/VR devices, thereby enabling direct 4D streaming for immersive applications.

% SUPP IMAGES HERE

\begin{figure*}[!ht]
\vspace{-6mm}
\centering
\includegraphics[width=5.5in]{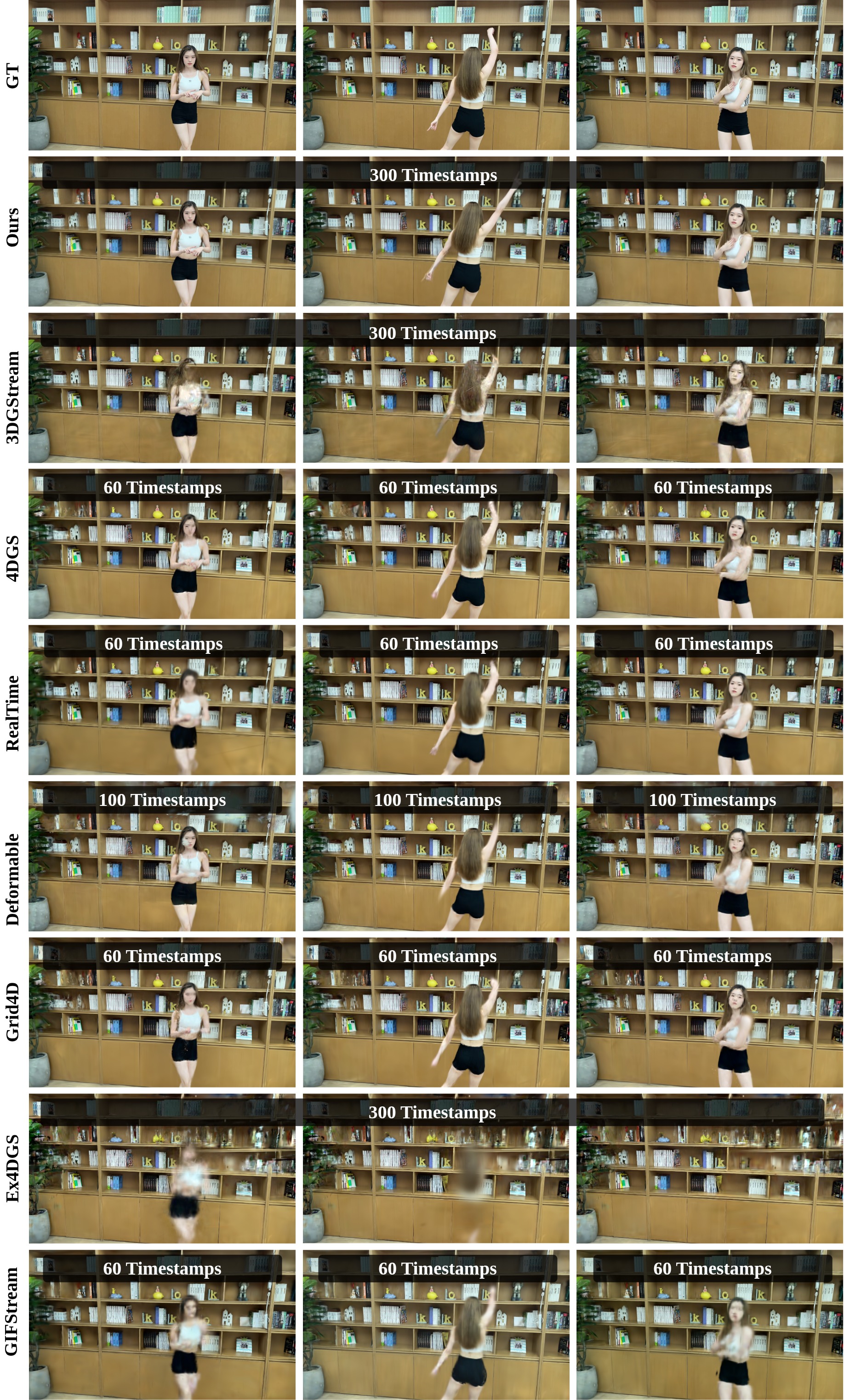} 
\caption{
Shows baseline comparison on SelfCap~\cite{xu2024longvolcap} dataset.
}
\label{fig:supp_baselines_selfcap}
\end{figure*}

\begin{figure*}[!ht]
\vspace{-5mm}
\centering
\includegraphics[width=6.3in]{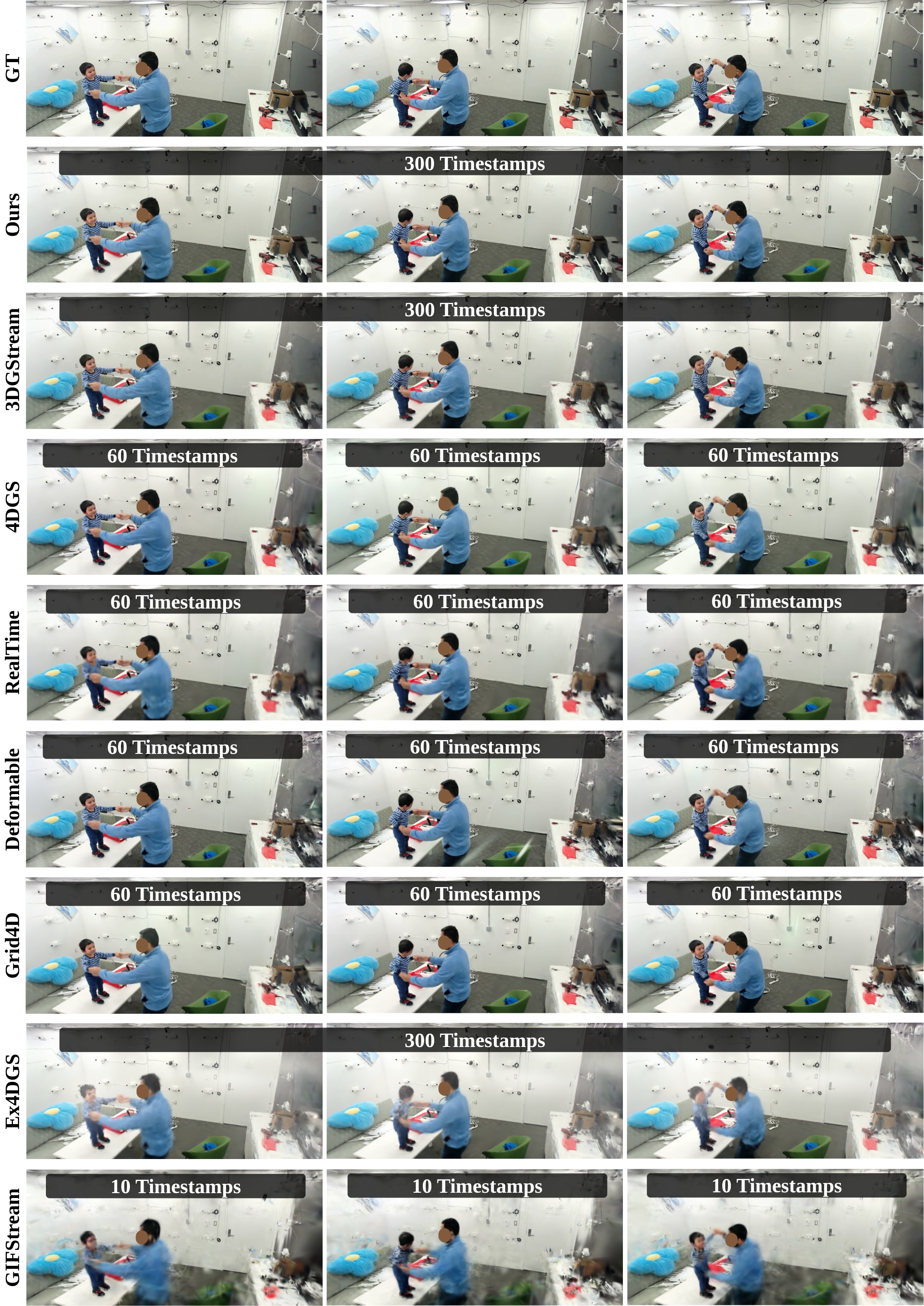} 
\caption{
Shows baselines comparison on \datasetname's \textit{Baby Dance} sequence.
}
\label{fig:supp_baby_result}
\end{figure*}

\begin{figure*}[!ht]
\centering
\includegraphics[width=6.8in]{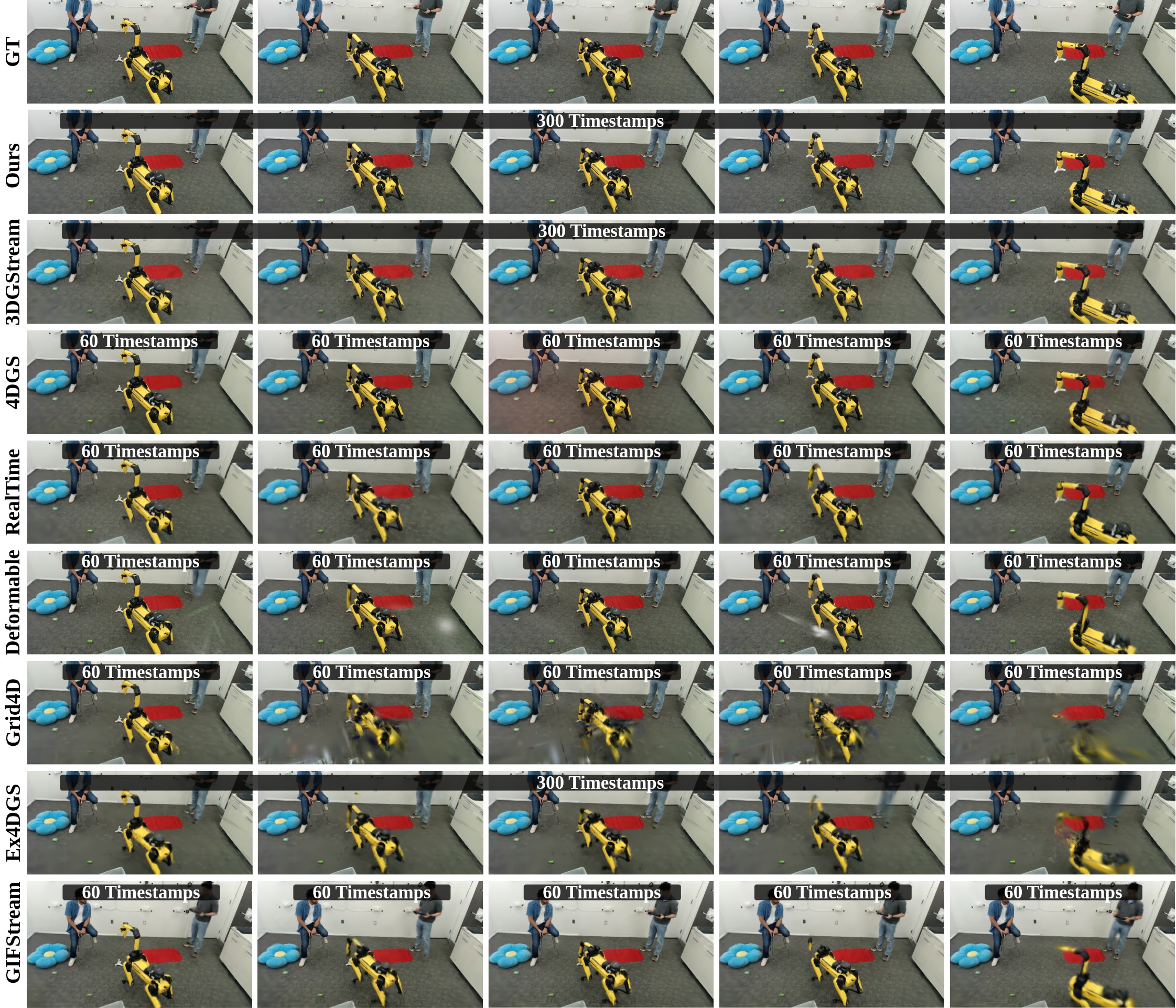} 
\caption{
Baselines comparison on \datasetname's \textit{SPOT} sequence.
}
\label{fig:supp_spot_result}
\end{figure*}


\end{document}